\documentclass{article}

\usepackage{dilab_arxiv}

\usepackage{subcaption}
\usepackage{amssymb}
\usepackage{mathtools}
\usepackage{amsthm}
\usepackage{tcolorbox}
\usepackage{array}
\usepackage{enumitem}
\usepackage{wrapfig}
\usepackage{algorithm}
\usepackage{algorithmic}
\usepackage{multirow}
\usepackage{colortbl}

\makeatletter
\@ifundefined{theHalgorithm}{}{}
\makeatother
\usepackage[capitalize,noabbrev]{cleveref}

\theoremstyle{plain}
\newtheorem{theorem}{Theorem}[section]

\theoremstyle{definition}
\newtheorem{definition}[theorem]{Definition}

\theoremstyle{remark}
\newtheorem{remark}[theorem]{Remark}

\usepackage[textsize=tiny]{todonotes}

\definecolor{impcolor}{RGB}{0,120,0}
\definecolor{degcolor}{RGB}{190,0,0}

\newcommand{\KL}{\mathrm{KL}}
\newcommand{\sg}{\operatorname{sg}}
\newcommand{\TV}{{\rm TV}}

\newcommand{\Acc}{\mathrm{Acc}}
\newcommand{\Harm}{\mathrm{Harm}}

\newcommand{\BC}{\mathrm{BothCorrect}}

\title{When Context Returns: \\ Toward Robust Internalization in On-Policy Distillation}

\runningtitle{Toward Robust Internalization in On-Policy Distillation}

\paperlogo{\includegraphics[height=1.5cm]{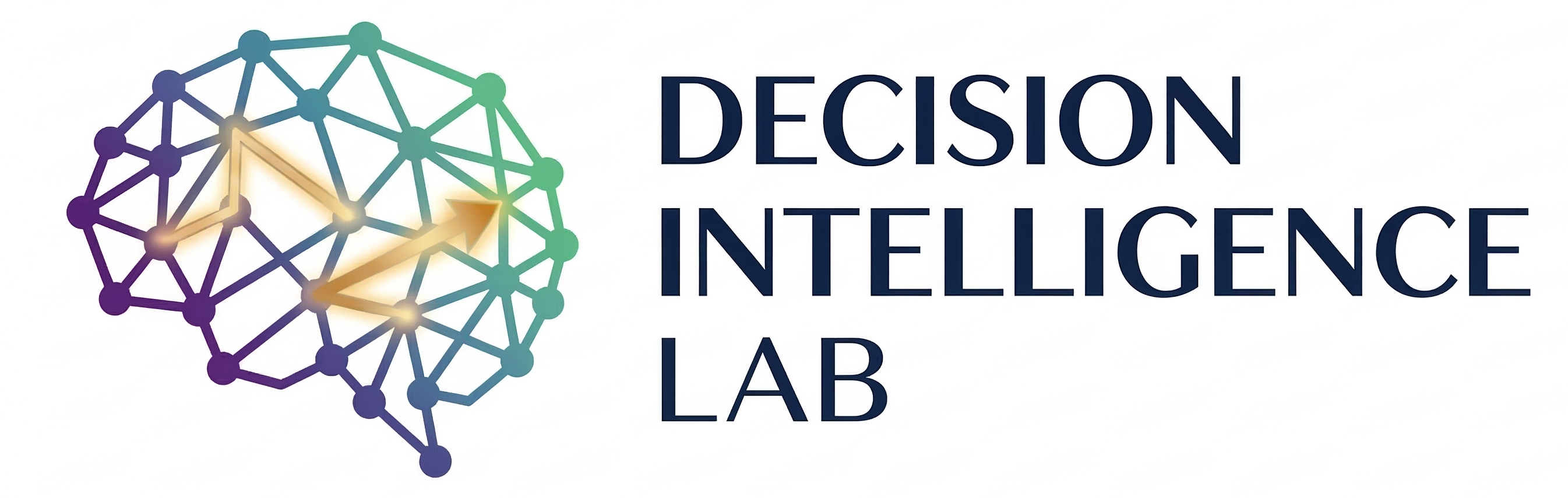}}

\author{
  Xun Wang$^{1}$, Ruishuo Chen$^{1}$, Zhuoran Li$^1$, Yu Chen$^1$, and Longbo Huang$^{1\,\text{\faEnvelope}}$%
  \\[0.3em] \normalfont 
  $^1$IIIS, Tsinghua University \quad 
  \text{\faEnvelope} Corresponding Author: longbohuang@tsinghua.edu.cn
}

\begin{document}

\maketitle
\thispagestyle{fancy}

\begin{abstract}
Recent work has shown that on-policy distillation can internalize privileged context, such as system prompts or task hints, into a student model so that the context is no longer needed at inference time. However, we identify a counterintuitive and previously unstudied phenomenon: reintroducing the original privileged context to the distilled student often degrades its performance, even on instances it already solves correctly without context. We term this phenomenon context-induced degradation and argue that robust internalization requires not only matching the teacher's context-conditioned behavior, but also remaining stable when the privileged context is reintroduced, a desirable property we call context invariance. To promote this property, we formulate a novel view-robust internalization risk and propose No-Context Anchoring (NCA), a lightweight yet effective consistency regularizer that uses the student's stop-gradient no-context output as an anchor and aligns its context-conditioned output via forward KL divergence. Across 14 configurations spanning diverse domains and model families, NCA improves context-conditioned accuracy in most settings and reduces context harm in 12 out of 14, while preserving or improving no-context performance, demonstrating greater robustness to context reintroduction.
\end{abstract}

\section{Introduction}
\label{sec:intro}

Language models are often augmented with privileged context, including system prompts encoding behavioral constraints, chain-of-thought traces providing expert reasoning~\citep{wei2022chain}, and game-state descriptions summarizing decision-relevant information.
Such context can improve performance through in-context learning~\citep{brown2020language,dong2022survey}, but retaining it at inference time increases latency and serving cost and may expose sensitive instructions to end users.
To avoid these drawbacks, on-policy self-distillation~\citep{zhao2026opsd,shenfeld2026self,hubotter2026reinforcement} trains a model to internalize privileged context by distilling its own context-conditioned behavior into the context-free behavior.
\citet{ye2026opcd} generalize this idea to on-policy context distillation by allowing distinct teacher and student models, enabling the student to perform well at deployment without requiring the context.

\begin{figure*}[t]
  \centering
  \includegraphics[width=\linewidth]{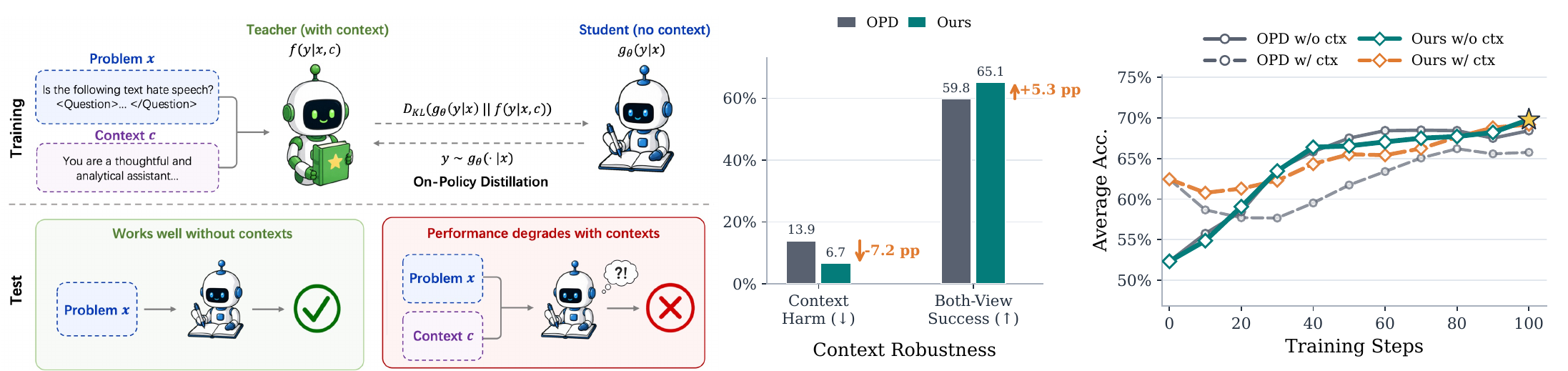}
  \caption{\textbf{Left}: OPD distills context-conditioned teacher behavior into a context-free student, yet reintroducing $c$ can reduce performance.
  \textbf{Middle}: Across all 14 configurations, NCA reduces context harm and improves both-view success, averaged over the final three checkpoints.
  \textbf{Right}: Three-checkpoint moving-average curves over the 12 no-thinking settings with a common 100-step schedule. OPD eventually performs worse with context, whereas NCA narrows the late-stage gap.}
  \label{fig:overview}
\end{figure*}

A natural question arises: if the student has truly internalized the privileged information, what happens when the context is \emph{reintroduced}? Intuitively, re-presenting already-internalized information shall be an idempotent operation~\citep{liu2026ider}: a student who has memorized the textbook shall not perform worse when allowed to consult it. Strikingly, we observe the opposite. Across the majority of settings, reintroducing the very context from which the student learned reduces its accuracy and can even overturn predictions that were correct without context (\Cref{fig:overview}, left). We call this counterintuitive phenomenon \emph{context-induced degradation}.

We trace this degradation to an asymmetry in the standard OPD objective. It optimizes exclusively for what we call \emph{privileged fidelity}, requiring the context-free student to match the context-conditioned teacher while leaving the student's behavior with context unconstrained. This limitation motivates a complementary requirement, \emph{context invariance}: once the privileged information has been internalized into the model's parameters, the student's output shall remain stable whether the context is present or absent.

To formalize these two requirements, we introduce a novel view-robust internalization risk that evaluates both the context-free and context-conditioned student views against the privileged teacher. From this risk, we derive a symmetric extension of standard OPD that separately aligns both student views with the teacher. Although context invariance follows at the exact optimum, we empirically find that this indirect teacher matching does not reliably realize it under approximate optimization.

Motivated by this gap, we formulate context invariance as an explicit constraint while preserving the standard OPD objective for privileged fidelity. We then propose \emph{No-Context Anchoring} (NCA), a lightweight yet effective penalty relaxation that treats the student's no-context output as a stop-gradient anchor and aligns the context-conditioned output to it through forward KL divergence, adding only one forward pass per training step.

Across $14$ configurations spanning diverse domains and model families, NCA improves context-conditioned accuracy in the majority of settings and reduces context harm in $12$ out of $14$, yielding an average $7.2$-percentage-point reduction in harm and a $5.3$-point gain in both-view success (\Cref{fig:overview}, middle). The training curves over the $12$ no-thinking settings further show that NCA substantially narrows the late-stage gap between context-free and context-conditioned accuracy (\Cref{fig:overview}, right).

In summary, our main contributions are:
\begin{itemize}[itemsep=2pt, topsep=2pt]
  \item We identify context-induced degradation, a counterintuitive phenomenon in which reintroducing privileged context can overturn correct context-free predictions, and introduce context invariance as a desirable property of robust internalization.
  \item We formulate robust internalization through a novel view-robust internalization risk that evaluates both student views against the privileged teacher, and propose No-Context Anchoring (NCA), a lightweight yet effective consistency regularizer that directly promotes context invariance using a stop-gradient no-context anchor and only one additional forward pass.
  \item Across 14 configurations spanning diverse domains and model families, NCA improves context-conditioned accuracy in most settings and reduces context harm in 12 out of 14, with an average reduction of 7.2 percentage points. A mechanistic case study further shows that NCA substantially increases the layer-wise hidden-state similarity between context-free and context-conditioned inputs.
\end{itemize}

\section{Related Work}
\label{sec:related}

\paragraph{Knowledge Distillation and On-Policy Distillation.}
Knowledge distillation~\citep{hinton2015distilling} transfers knowledge from a teacher to a student by matching their output distributions, and has been widely applied to LLMs for distilling reasoning capabilities~\citep{DeepSeekAI2025DeepSeekR1IR}, instruction following~\citep{ouyang2022training}, and general capabilities~\citep{gu2024minillm,ko2024distillm,xu2024survey}. Recently, on-policy distillation (OPD)~\citep{agarwal2024onpolicy,deepseekai2026deepseekv4,xiao2026mimo,fu2026revisiting,li2026rethinking,jin2026entropyaware} generates training data from the student's own distribution, mitigating the train--inference distribution mismatch of off-policy approaches. When the teacher is the model itself conditioned on privileged context, this becomes on-policy self-distillation~\citep{zhao2026opsd,shenfeld2026self,hubotter2026reinforcement,penaloza2026privileged}. \citet{ye2026opcd} further generalize this setting by allowing the context-conditioned teacher and the student to be instantiated by models of different sizes. \citet{zhao2026ophsd} extend this idea by treating the reasoning harness itself as privileged context to be distilled. For a broader discussion of OPD and its variants, we refer readers to \citet{song2026survey}.

\paragraph{Analysis and Extensions of On-Policy Distillation.}
A growing body of work analyzes the mechanisms underlying on-policy distillation. For example, \citet{li2026rethinking} identify conditions governing its success and failure, while \citet{fu2026revisiting} catalog three failure modes of sampled-token OPD. \citet{kaur2026rethinking} study how privileged-context distillation affects the context-free performance of thinking models, identifying performance degradation under long reasoning budgets. Complementary to these analyses, several recent works combine on-policy self-distillation with reinforcement learning from verifiable rewards~\citep{yang2026rlsd,li2025srpo} to further improve training efficiency and reasoning performance. Our focus is orthogonal to both lines of work: rather than explaining when OPD succeeds, diagnosing its training failures, or improving its optimization signal, we study what happens after privileged context has been successfully internalized and is subsequently reintroduced. We show that privileged fidelity alone is insufficient for robust internalization without explicitly promoting context invariance.

\paragraph{Consistency Regularization.}
Consistency regularization encourages a model to produce similar outputs across different input views that shall preserve the same target behavior~\citep{berthelot2019mixmatch,sohn2020fixmatch}. Recent applications include robustness to paraphrased instructions~\citep{zhao2024improving}, self-rewarding alignment~\citep{wang2025cream} in LLMs, and idempotent knowledge retention in continual learning~\citep{liu2026ider}. These works enforce consistency across diverse perturbations and aspects of model behavior. In contrast, we apply this principle to the context-free and context-conditioned views of a distilled student, directly promoting stability under context reintroduction.

\section{Problem Formulation}
\label{sec:problem}

Consider a student model $g_\theta$ parameterized by $\theta$, an input $x$, and a fixed privileged context $c$ (e.g., a system prompt, task hint, or game-state scaffold).\footnote{Because $c$ is given and fixed for each task, the KL in \Cref{eq:opcd} contains no irreducible entropy term~\citep{yang2026rlsd}. Yet even in this favorable setting, optimizing privileged fidelity alone does not guarantee context invariance.}
We denote the two student views as $q_x \triangleq g_\theta(y \mid x)$ and $q_{x,c} \triangleq g_\theta(y \mid x, c)$.
On-policy distillation with privileged context~\citep{ye2026opcd} trains the student by minimizing:
\begin{equation}
  \mathcal{L}_{\mathrm{OPD}} = \mathbb{E}_{x \sim \mathcal{D}} \left[ \KL\!\left(q_x \,\middle\|\, f(y \mid x, c)\right) \right],
  \label{eq:opcd}
\end{equation}
where $f(y \mid x, c)$ is the teacher distribution. In practice, the KL is computed at the token level and summed over the sequence.

\begin{wrapfigure}{r}{0.40\textwidth}
  \vspace{-12pt}
  \centering
  \includegraphics[width=0.32\textwidth]{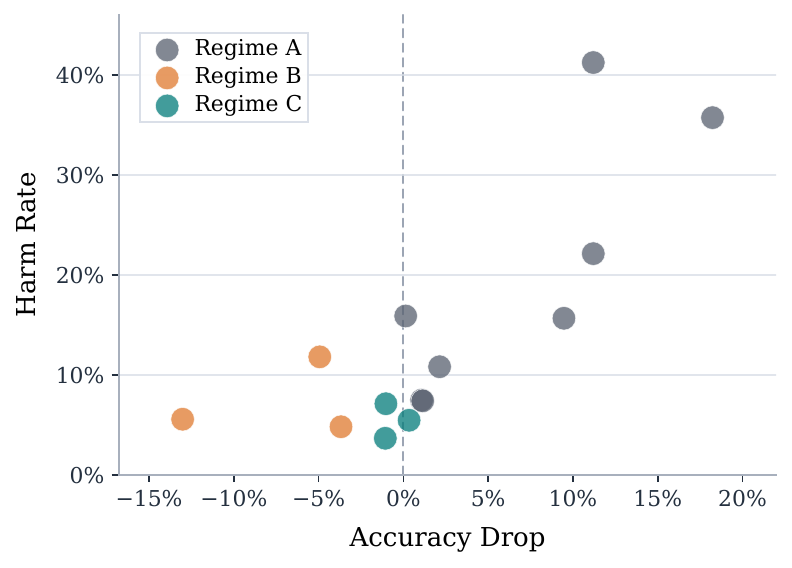}
  \caption{Regimes of context reintroduction after OPD. x-axis: accuracy drop upon context reintroduction; y-axis: $\Harm$ rate.}
  \label{fig:regimes}
  \vspace{-12pt}
\end{wrapfigure}
Note that \Cref{eq:opcd} only constrains $q_x$. The context-conditioned view $q_{x,c}$ receives no learning signal and is not directly constrained.
We argue that a necessary property for successful internalization is \emph{context invariance}: if the model has truly absorbed $c$ into its parameters, conditioning on $c$ becomes redundant and the two views shall agree, i.e., $q_x \approx q_{x,c}$. Although one might expect a sufficiently expressive model to satisfy this property after training, we find that this is not the case in the majority of settings.

\paragraph{Three regimes of context interaction.}
\label{sec:regimes}
We evaluate OPD across 14 configurations spanning various model families and three task domains: system-prompt following, Text Games, and Mathematical Reasoning. \Cref{fig:regimes} plots each configuration by the accuracy drop induced by context reintroduction, $\Acc_x-\Acc_{x,c}$, against context harm, $\Harm=P(q_{x,c}\text{ wrong}\mid q_x\text{ correct})$, which measures how often context overturns a correct no-context prediction. Three regimes emerge. In \textbf{Regime A} (8/14 configurations), context acts as a harmful perturbation: although the student already performs well without it, reintroducing it causes both an accuracy drop and a high harm rate. In \textbf{Regime B}, context remains a useful scaffold, suggesting incomplete internalization. In \textbf{Regime C}, the accuracy gap between the two views is small and context harm is minimal. Crucially, Regime~A is the dominant pattern, covering more than half of all configurations. This shows that context-induced degradation is not an isolated failure case, but a recurring limitation of standard OPD, motivating the explicit promotion of context invariance.

\section{Method: No-Context Anchoring}
\label{sec:method}
In this section, we first formalize a novel objective, view-robust internalization risk, that captures both privileged fidelity and context invariance. We then show that exact minimization of this risk implies context invariance, whereas indirect teacher matching may not reliably realize it under approximate optimization. Finally, we propose No-Context Anchoring (NCA), a regularizer that directly promotes cross-view consistency while preserving the OPD objective.
The full procedure is summarized in \Cref{alg:nca}.
\subsection{View-Robust Internalization Risk}
\label{sec:theory_risk}
First, we introduce the view-robust internalization risk. Let $x$ be the input, $y$ the output, and $c$ the privileged context.
Let $T_c \triangleq f(y\mid x,c)$ denote the teacher conditioned on~$c$.
At deployment, the student may or may not receive the context.
We model this with a Bernoulli view variable $V\sim\mathrm{Bernoulli}(p_v)\triangleq\pi_V$, $p_v\in(0,1)$, and define the two student views
\begin{equation}
Q_\theta(y\mid x,v) \triangleq \begin{cases}
g_\theta(y\mid x) & v=0,\\
g_\theta(y\mid x,c) & v=1.
\end{cases}
\end{equation}
Then the joint distribution of the view and output is $P_\theta(v,y|x)=\pi_V(v)Q_\theta(y|x,v)$.
For notational convenience, write $q_x \triangleq g_\theta(\cdot\mid x)$ and $q_{x,c} \triangleq g_\theta(\cdot\mid x,c)$.

If the student has truly internalized the privileged information, its output shall not depend on whether $c$ is present.
The ideal joint distribution is therefore $P_T(v,y\mid x) = \pi_V(v)\,T_c(y)$, and we measure the gap via

\begin{definition}[View-robust internalization risk]
\begin{equation}
\mathcal R(\theta;p_v) \triangleq
\mathbb E_{x\sim\mathcal D}\,
\KL\!\left(
P_\theta(V,Y\mid x)
\,\middle\|\,
P_T(V,Y\mid x)
\right).
\label{eq:view_robust_risk}
\end{equation}
\end{definition}

Expanding \Cref{eq:view_robust_risk} yields
\begin{equation}
\mathcal R(\theta;p_v)
=
(1-p_v)\,\mathbb E_{x \sim \mathcal{D}}\,\KL(q_x\|T_c)
+
p_v\,\mathbb E_{x \sim \mathcal{D}}\,\KL(q_{x,c}\|T_c).
\label{eq:risk_expansion}
\end{equation}

The first term is precisely the standard OPD objective.
The full risk adds a second, symmetric term that requires the context-conditioned view to also match the teacher. We refer to this joint objective as \textbf{DualOPD}. Then, any parameter $\theta^*$ that achieves $\mathcal R(\theta^*;p_v)=0$ satisfies $q_x = q_{x,c} = T_c$ for $\mathcal D$-almost every $x$. Therefore, context invariance (i.e., $q_x = q_{x,c}$) is a necessary condition for the view-robust optimum.

\subsection{The Need for Direct Invariance Optimization}
\label{sec:theory_dualopd}

Although DualOPD directly minimizes $\mathcal R$, context invariance is a consequence of \emph{exact} minimization but is not explicitly enforced during optimization.
In practice, with finite model capacity, limited optimization steps, and top-$k$ logit truncation, the two teacher-matching terms may each reach small values without guaranteeing that $q_x$ and $q_{x,c}$ converge to the same solution.
In particular, when $T_c$ is multi-modal, each view can approximate $T_c$ by concentrating on different modes, both achieving small divergence from the teacher while remaining far apart from each other.

Our ablation study (Section \ref{sec:ablation}) confirms this empirically: DualOPD suffers a severe early $\Acc_{x,c}$ crash and learns $\Acc_x$ most slowly among all methods, consistent with the two views diverging under approximate optimization despite both tracking the teacher.
This motivates a formulation that directly promotes context invariance rather than relying on it to emerge as a side effect of exact teacher matching.

\subsection{No-Context Anchoring}
Therefore, we preserve the OPD objective for privileged fidelity while adding an explicit context-invariance constraint:
\begin{equation}
  \mathcal{P}:\quad \min_\theta \; \mathcal{L}_{\mathrm{OPD}}(\theta) \quad \text{s.t. } \mathbb{E}_{x\sim\mathcal D}
\left[D(q_x,q_{x,c})\right] \leq \varepsilon,
  \label{eq:constrained}
\end{equation}
where $D$ is a divergence between the two views.
In practice, we relax this into a penalized form with a fixed weight $\beta > 0$:
\begin{equation}
  \mathcal{L} = \mathcal{L}_{\mathrm{OPD}} + \beta \, \mathbb{E}_{x \sim \mathcal{D}}\!\left[D(q_x,q_{x,c})\right].
  \label{eq:full}
\end{equation}

The divergence $D$ in \Cref{eq:full} can take many forms.
We choose the forward KL divergence $D = \KL(q_x \| q_{x,c})$, which is mode-covering and thus prevents $q_{x,c}$ from collapsing modes of $q_x$.
An additional benefit is that the expectation falls under $q_x$, allowing us to reuse the on-policy samples already generated by OPD without a separate rollout from $q_{x,c}$.
To further avoid the gradient of this term interfering with the OPD objective through $q_x$, we apply a stop-gradient to anchor the no-context view.
We call the resulting regularizer \emph{No-Context Anchoring} (NCA):
\begin{equation}
  \mathcal{L}_{\mathrm{NCA}} \triangleq \KL\!\left(\sg\!\left[g_\theta(\cdot \mid x)\right] \,\middle\|\, g_\theta(\cdot \mid x, c)\right),
  \label{eq:nca}
\end{equation}
where $\sg[\cdot]$ denotes the stop-gradient operator.

The stop-gradient prevents $\mathcal{L}_{\mathrm{NCA}}$ from backpropagating through the $q_x$ branch, preserving the no-context output as a reference within each update.
Minimizing the NCA regularizer directly penalizes cross-view disagreement: $\KL(q_x\|q_{x,c})=0$ if and only if $q_x = q_{x,c}$.
In contrast, DualOPD promotes context invariance only indirectly through separate teacher matching, which, as discussed in \Cref{sec:theory_dualopd}, can leave $q_x$ and $q_{x,c}$ far apart under approximate optimization.
We compare against alternative divergence choices in our design analysis (\Cref{sec:ablation}).
\subsection{Bound Analysis}

The NCA regularizer provides a quantitative connection to context-induced behavior change. For any task-relevant event $A_x$ associated with input $x$ (e.g., answering correctly), NCA upper-bounds the probability change caused by context reintroduction. Concretely, by the definition of total variation, Pinsker's inequality, and Jensen's inequality:
$$\mathbb E_{x \sim \mathcal{D}} |q_{x,c}(A_x) - q_x(A_x)|\leq \mathbb E_{x \sim \mathcal{D}}\left[\TV(q_{x,c},q_x)\right] \leq \mathbb E_{x \sim \mathcal{D}}\left[\sqrt{\tfrac{1}{2}\KL(q_x\|q_{x,c})}\right] \leq \sqrt{\tfrac{1}{2}\,\mathbb E_{x \sim \mathcal{D}}\left[\KL(q_x\|q_{x,c})\right]} \leq \sqrt{\tfrac{\varepsilon}{2}}.$$

\begin{remark}
If there exists $\theta^*$ such that $g_{\theta^*}(\cdot\mid x) = g_{\theta^*}(\cdot\mid x,c) = T_c$ for $\mathcal D$-a.e.\ $x$, both DualOPD and NCA achieve the zero-loss solution. That is, NCA does not sacrifice the expressiveness of the view-robust objective. The practical difference is that DualOPD relies on context invariance emerging as a side effect of exact teacher matching, whereas NCA directly enforces it, making it more robust under finite optimization.
\end{remark}
\subsection{Computational Cost}

The only additional cost of NCA is one forward pass per training step for $q_{x,c} = g_\theta(\cdot \mid x, c)$. Specifically, the $\sg[q_x]$ logits are detached from the forward pass that OPD already performs, and the on-policy sequences from OPD are reused to evaluate $\mathcal{L}_{\mathrm{NCA}}$ without a separate rollout.

\begin{algorithm}[!ht]
\caption{OPD with No-Context Anchoring (NCA)}
\label{alg:nca}
\begin{algorithmic}[1]
\REQUIRE Dataset $\mathcal{D}$, student $g_\theta$, teacher $f$, context $c$, weight $\beta$
\FOR{each batch $\{x_i\} \sim \mathcal{D}$}
  \STATE Sample $y_i \sim g_\theta(\cdot \mid x_i)$ \hfill (on-policy rollout)
  \STATE $\mathcal{L}_{\mathrm{OPD}} \leftarrow \frac{1}{|\mathcal{B}|}\sum_i \sum_t \KL\!\left(g_\theta(\cdot \mid x_i, y_{i,<t}) \,\middle\|\, f(\cdot \mid x_i, c, y_{i,<t})\right)$
  \STATE $\mathcal{L}_{\mathrm{NCA}} \leftarrow \frac{1}{|\mathcal{B}|}\sum_i \sum_t \KL\!\left(\sg\!\left[g_\theta(\cdot \mid x_i, y_{i,<t})\right] \,\middle\|\, g_\theta(\cdot \mid x_i, c, y_{i,<t})\right)$
  \STATE Update $\theta$ via $\nabla_\theta (\mathcal{L}_{\mathrm{OPD}} + \beta \cdot \mathcal{L}_{\mathrm{NCA}})$
\ENDFOR
\end{algorithmic}
\end{algorithm}

\section{Experiments}
\label{sec:experiments}

We evaluate No-Context Anchoring (NCA) across 14 configurations spanning three domains (System-Prompt QA, Text Games, and Mathematical reasoning), five tasks, and six models, aiming to answer:
(i) How prevalent and severe is context-induced degradation across tasks and model families, and can NCA mitigate it without sacrificing no-context performance?
(ii) How do the choice of divergence and the regularization strength $\beta$ affect NCA's performance?
(iii) What mechanisms underlie NCA's effect on the model's internal representations?
(iv) How robust is NCA across model scales, cross-domain shifts, and variable privileged contexts?
\subsection{Experimental Setup}
\label{sec:exp_setup}

\textbf{Tasks.}
Our experiments cover three types of tasks.
\textbf{\emph{System-Prompt QA}:} the privileged context is a task-specific system prompt available to the teacher but intended to be internalized by the student.
We use MedMCQA~\citep{pal2022medmcqa}, a medical multiple-choice QA dataset with 500 test samples, and a Safety dataset combining TweetEval~\citep{tweeteval}, HateCheck~\citep{hatecheck}, and ETHOS~\citep{ethos} for harmful-content classification, with 498 test samples.
\textbf{\emph{Text Games}:} the privileged context is a game-state scaffold, such as hints about the optimal move.
We use Sokoban ($6\times6$ grid, one box; spatial reasoning) and FrozenLake ($3\times3$ grid, two holes; navigation) from TextArena~\citep{textarena}, both with a maximum of five game steps.
\textbf{\emph{Mathematical reasoning}:} the privileged context provides general problem-solving guidance, such as first identifying the problem type and planning an appropriate solution strategy. We use DAPO-Math-17K \cite{dapo} with $1,000$ test samples.
All training and test datasets, data splits, and text-game configurations follow \citet{ye2026opcd}.

\textbf{Models.}
For System-Prompt QA, we evaluate four models: Llama-3.1-8B-Instruct~\citep{grattafiori2024llama}, Llama-3.2-3B-Instruct~\citep{grattafiori2024llama}, Qwen2.5-7B-Instruct~\citep{yang2024qwen2}, and Qwen3-8B~\citep{yang2025qwen3} (no-thinking mode), each serving as both teacher and student.
For Text Games, Qwen3-4B-Instruct-2507~\citep{yang2025qwen3} serves as the teacher, with the same model as the student for self-distillation and Qwen3-1.7B~\citep{yang2025qwen3} (no-thinking mode) for cross-model distillation.
For Mathematical Reasoning, we use Qwen3-1.7B and Qwen3-8B with thinking mode, each serving as both teacher and student.

\textbf{Training.}
We use full-parameter fine-tuning with a learning rate of $5 \times 10^{-6}$ for 100 steps on System-Prompt QA and Text Games, and 90 steps on Mathematical Reasoning due to its smaller training set.
The batch size for System-Prompt QA and Mathematical Reasoning is 128, and for Text Games is 64.
We set $\beta = 0.5$ for all settings except Qwen3-4B-Instruct-2507$\to$Qwen3-1.7B in Text Games where we use $\beta = 0.1$.
For System-Prompt QA, we use MetaSPO~\citep{wan2024metaspo} to optimize suitable system prompts for each base model.
For Text Games and Mathematical Reasoning, we construct the privileged context using the experience templates from \citet{ye2026opcd}: scenario-specific gameplay guidance for each text game and general problem-solving guidance for math. The system prompts and experiences are provided in Appendix~\ref{apd:implementation}.
The maximum response length is $512$ tokens for System-Prompt QA, $1024$ tokens for Text Games, and $16,384$ tokens for Mathematical Reasoning.

\textbf{Evaluation.}
We evaluate every 10 training steps under both views (no context and context reintroduced) using the following metrics:
\begin{itemize}[itemsep=1pt, topsep=2pt]
  \item $\Acc_x$ / $\mathrm{WR}_x$ and $\Acc_{x,c}$ / $\mathrm{WR}_{x,c}$: task performance (accuracy for QA, win rate for games) under the no-context and context-reintroduced views, respectively.
  \item $\Harm$: conditional probability of context flipping a correct prediction to incorrect, i.e., $P(q_{x,c} \text{ wrong} \mid q_x \text{ correct})$.
  \item $\BC$ / $\mathrm{BothWon}$: fraction of instances correct (or games won) under both views.
\end{itemize}
For System-Prompt QA, we use greedy decoding and evaluate on the full test set.
For Text Games, we evaluate on 128 game instances, each with a single stochastic rollout ($\tau{=}0.7$, top-$p{=}0.8$, top-$k{=}20$).
For Mathematical Reasoning, we evaluate with $20$ sampled response per test sample ($\tau{=}0.6$, top-$p{=}0.95$, top-$k{=}20$).
We report the average performance over the final three checkpoints.
Throughout the experiments, we use OPD to refer specifically to on-policy distillation with privileged context. Our baseline is implemented based on the codebase of \citet{ye2026opcd}.

\textbf{Implementation.}
Our code is built on top of the codebase of \citet{ye2026opcd}, using veRL~\citep{sheng2024verl} as the training engine with FSDP for distributed training and vLLM~\citep{kwon2023efficient} for on-policy rollout generation.
All KL divergences are computed over the top-$k{=}256$ logits with renormalization to prevent reward hacking from low-probability tokens. 
All experiments are conducted on a single server equipped with 8$\times$NVIDIA H100 GPUs.

\subsection{Evaluation Results}
\label{sec:results}

\begin{figure}[!ht]
  \centering
  \includegraphics[width=0.6\linewidth]{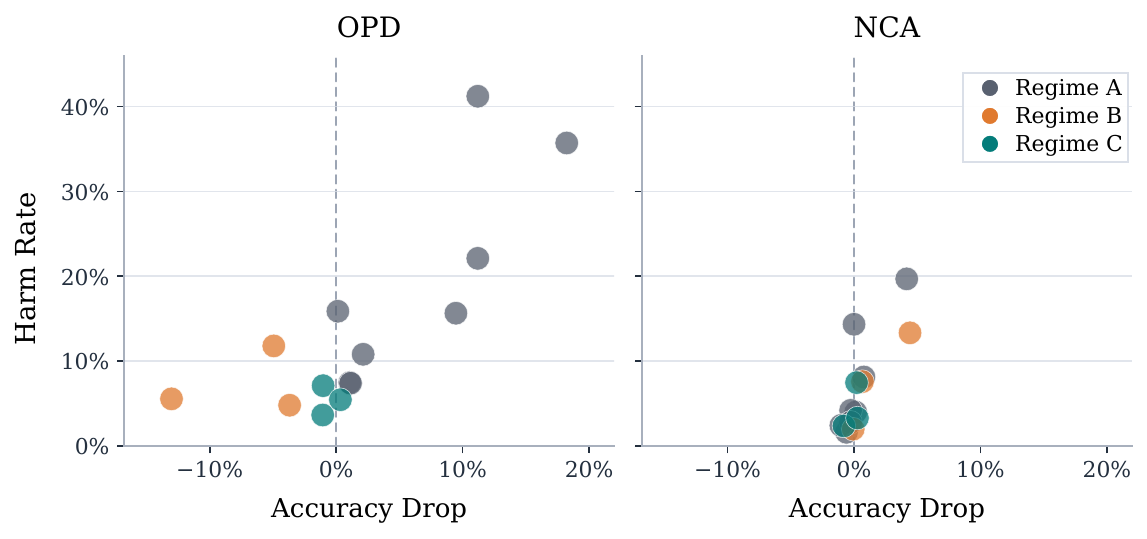}
  \caption{Overall effect of NCA across all 14 settings. Left: OPD. Right: NCA. Each point is one (model, task) setting, colored by regime. NCA moves most points toward the origin, reducing both accuracy drop and harm rate.}
  \label{fig:overall_effect}
\end{figure}

\Cref{fig:overall_effect} provides a bird's-eye view of NCA's effect across all 14 settings. Under OPD (left), Regime~A points are scattered in the high-harm, high-accuracy-drop region. After applying NCA (right), most points move toward the origin, indicating that NCA broadly reduces context-induced degradation across diverse models and tasks.

\subsubsection{System-Prompt QA}
\label{sec:sys_results}

\begin{table*}[!ht]
\centering
\caption{System-Prompt QA results on MedMCQA and Safety. All metrics are reported in \%. The better result between OPD and NCA is shown in \textbf{bold} for each metric. Values are averaged over the final three checkpoints.}
\setlength{\tabcolsep}{3.5pt}
\begin{small}
\begin{tabular}{ll cccc cccc}
\toprule
& & \multicolumn{4}{c}{\textbf{MedMCQA}} & \multicolumn{4}{c}{\textbf{Safety}} \\
\cmidrule(lr){3-6} \cmidrule(lr){7-10}
\textbf{Model} & \textbf{Method} & $\Acc_x\!\uparrow$ & $\Acc_{x,c}\!\uparrow$ & $\Harm\!\downarrow$ & $\BC\!\uparrow$ & $\Acc_x\!\uparrow$ & $\Acc_{x,c}\!\uparrow$ & $\Harm\!\downarrow$ & $\BC\!\uparrow$ \\
\midrule
\multirow{3}{*}{Llama-3.1-8B-Ins}
 & \textcolor{gray}{Base} & \textcolor{gray}{65.8} & \textcolor{gray}{79.2} & \textcolor{gray}{--} & \textcolor{gray}{--} & \textcolor{gray}{74.1} & \textcolor{gray}{77.5} & \textcolor{gray}{--} & \textcolor{gray}{--} \\
 & OPD & 77.1 & 75.0 & 10.8 & 68.8 & 75.3 & 74.2 & 7.5 & 69.7 \\
 & NCA (Ours) & \textbf{78.0} & \textbf{77.9} & \textbf{4.0} & \textbf{74.9} & \textbf{76.4} & \textbf{76.6} & \textbf{4.2} & \textbf{73.2} \\
\midrule
\multirow{3}{*}{Llama-3.2-3B-Ins}
 & \textcolor{gray}{Base} & \textcolor{gray}{60.0} & \textcolor{gray}{73.2} & \textcolor{gray}{--} & \textcolor{gray}{--} & \textcolor{gray}{60.8} & \textcolor{gray}{74.1} & \textcolor{gray}{--} & \textcolor{gray}{--} \\
 & OPD & \textbf{74.8} & 65.3 & 15.7 & 63.1 & 73.8 & \textbf{77.5} & 4.8 & 70.3 \\
 & NCA (Ours) & 74.7 & \textbf{75.3} & \textbf{1.6} & \textbf{73.5} & \textbf{74.4} & 74.4 & \textbf{2.0} & \textbf{72.9} \\
\midrule
\multirow{3}{*}{Qwen2.5-7B-Ins}
 & \textcolor{gray}{Base} & \textcolor{gray}{51.4} & \textcolor{gray}{61.4} & \textcolor{gray}{--} & \textcolor{gray}{--} & \textcolor{gray}{66.9} & \textcolor{gray}{75.7} & \textcolor{gray}{--} & \textcolor{gray}{--} \\
 & OPD & 58.9 & \textbf{63.8} & 11.8 & 51.9 & 73.3 & 72.2 & 7.4 & 67.9 \\
 & NCA (Ours) & \textbf{62.5} & 61.8 & \textbf{7.6} & \textbf{57.7} & \textbf{74.5} & \textbf{74.8} & \textbf{2.8} & \textbf{72.4} \\
\midrule
\multirow{3}{*}{Qwen3-8B$^\dagger$}
 & \textcolor{gray}{Base} & \textcolor{gray}{71.2} & \textcolor{gray}{71.4} & \textcolor{gray}{--} & \textcolor{gray}{--} & \textcolor{gray}{79.3} & \textcolor{gray}{79.9} & \textcolor{gray}{--} & \textcolor{gray}{--} \\
 & OPD & 69.3 & 70.4 & 3.7 & 66.8 & 78.5 & 78.2 & 5.5 & 74.2 \\
 & NCA (Ours) & \textbf{69.9} & \textbf{70.7} & \textbf{2.4} & \textbf{68.3} & \textbf{79.2} & \textbf{78.9} & \textbf{3.3} & \textbf{76.6} \\
\bottomrule
\end{tabular}
\end{small}
\vspace{2pt}
\raggedright{\scriptsize $^\dagger$No-thinking mode.}
\label{tab:sys_prompt}
\end{table*}
\Cref{tab:sys_prompt} reveals two main findings.

\textbf{Context-induced degradation is widespread.}
Under standard OPD, all eight settings exhibit non-trivial context harm.
The most severe case is Llama-3.2-3B on MedMCQA, where $\Harm$ reaches 15.7\% and $\Acc_{x,c}$ falls 9.5 points below $\Acc_x$.
Even Qwen3-8B, whose base model performs similarly under the two views, exhibits a non-trivial harm rate of 5.5\% after OPD on Safety, showing that strong initial performance alone does not ensure robustness to context reintroduction.

\textbf{NCA consistently improves robustness.}
NCA reduces $\Harm$ and improves $\BC$ in all eight settings, while also improving $\Acc_x$ in seven of eight.
The most pronounced improvement occurs on Llama-3.2-3B MedMCQA, where $\Harm$ decreases by 14.1 percentage points and $\BC$ increases by 10.4 points.
Notably, although NCA directly regularizes the context-conditioned view, the frequent gains in $\Acc_x$ suggest that promoting cross-view consistency may also facilitate better internalization.
The only notable trade-off appears on Qwen2.5-7B MedMCQA, a Regime~B setting in which context remains beneficial: NCA lowers $\Acc_{x,c}$ by 2.0 points, but still improves $\Acc_x$ by 3.6 points and reduces $\Harm$ by 4.2 points.
This suggests that NCA can remain beneficial when internalization is incomplete, although stronger cross-view consistency may come at some cost to context-conditioned performance.

\subsubsection{Text Games}
\label{sec:game_results}

\begin{table*}[t]
\centering
\caption{Text Games results on Sokoban and FrozenLake. All metrics are reported in \%. The better result between OPD and NCA is shown in \textbf{bold} for each metric. Values are averaged over the final three checkpoints.}
\setlength{\tabcolsep}{4pt}
\begin{small}
\begin{tabular}{ll cccc cccc}
\toprule
& & \multicolumn{4}{c}{\textbf{Sokoban}} & \multicolumn{4}{c}{\textbf{FrozenLake}} \\
\cmidrule(lr){3-6} \cmidrule(lr){7-10}
\textbf{Model} & \textbf{Method} & WR$_x\!\uparrow$ & WR$_{x,c}\!\uparrow$ & $\Harm\!\downarrow$ & $\mathrm{BothWon}\!\uparrow$ & WR$_x\!\uparrow$ & WR$_{x,c}\!\uparrow$ & $\Harm\!\downarrow$ & $\mathrm{BothWon}\!\uparrow$ \\
\midrule
\multirow{3}{*}{Qwen3-4B-Ins (self)}
 & \textcolor{gray}{Base} & \textcolor{gray}{5.5} & \textcolor{gray}{43.0} & \textcolor{gray}{--} & \textcolor{gray}{--} & \textcolor{gray}{65.6} & \textcolor{gray}{75.8} & \textcolor{gray}{--} & \textcolor{gray}{--} \\
 & OPD & \textbf{51.8} & 40.6 & 22.1 & 40.4 & 63.8 & 52.6 & 41.2 & 37.8 \\
 & NCA (Ours) & 51.3 & \textbf{50.5} & \textbf{8.1} & \textbf{47.1} & \textbf{69.5} & \textbf{65.4} & \textbf{19.7} & \textbf{53.4} \\
\midrule
\multirow{3}{*}{Qwen3-4B-Ins$\to$1.7B$^\dagger$}
 & \textcolor{gray}{Base} & \textcolor{gray}{27.3} & \textcolor{gray}{27.3} & \textcolor{gray}{--} & \textcolor{gray}{--} & \textcolor{gray}{0.0} & \textcolor{gray}{10.9} & \textcolor{gray}{--} & \textcolor{gray}{--} \\
 & OPD & 51.0 & 32.8 & 35.7 & 32.8 & 73.4 & \textbf{86.5} & \textbf{5.6} & \textbf{69.3} \\
 & NCA (Ours) & \textbf{52.6} & \textbf{53.6} & \textbf{2.4} & \textbf{51.3} & \textbf{74.5} & 70.1 & 13.3 & 64.3 \\
\bottomrule
\end{tabular}
\end{small}
\vspace{2pt}
\raggedright{\scriptsize $^\dagger$No-thinking mode.}
\label{tab:textgame}
\end{table*}

\Cref{tab:textgame} reports the Text Games results, where context-induced degradation is often more pronounced than in System-Prompt QA due to the sequential nature of the tasks.

\textbf{NCA improves robustness in most settings.}
Under standard OPD, three of the four settings exhibit substantial context harm, reaching 41.2\% on self-distilled FrozenLake and 35.7\% on cross-model Sokoban.
NCA markedly reduces $\Harm$ and improves $\mathrm{BothWon}$ in all three cases.
In particular, on cross-model Sokoban, NCA reduces $\Harm$ from 35.7\% to 2.4\% while improving both $\mathrm{WR}_x$ and $\mathrm{WR}_{x,c}$.
These results show that NCA generally remains effective under both self-distillation and cross-model distillation in sequential decision-making tasks.

\textbf{Boundary case.}
Cross-model FrozenLake is the only setting in which NCA increases $\Harm$.
This is also a Regime~B case that retains a strong benefit from context under OPD, with $\mathrm{WR}_{x,c}=86.5\%$ compared with $\mathrm{WR}_x=73.4\%$.
Anchoring the context-conditioned view to the no-context output reduces this benefit, lowering $\mathrm{WR}_{x,c}$ by 16.4 points while improving $\mathrm{WR}_x$ by only 1.1 points.
This observation motivates a promising future direction: adapting the anchor to the degree of internalization rather than always fixing it to the no-context view. We examine the corresponding training dynamics in \Cref{fig:tdb}.

\subsubsection{Mathematical Reasoning}
\label{sec:math_results}
\begin{table*}[!ht]
\centering
\caption{Mathematical Reasoning results. All metrics are reported in \%. The better result between OPD and NCA is shown in \textbf{bold} for each metric. Values are averaged over the final three checkpoints.}
\setlength{\tabcolsep}{3.5pt}
\begin{small}
\begin{tabular}{ll cccc}
\toprule
\textbf{Model} & \textbf{Method} & $\Acc_x\!\uparrow$ & $\Acc_{x,c}\!\uparrow$ & $\Harm\!\downarrow$ & $\BC\!\uparrow$ \\
\midrule
\multirow{3}{*}{Qwen3-1.7B}
 & \textcolor{gray}{Base} & \textcolor{gray}{60.6} & \textcolor{gray}{63.4} & \textcolor{gray}{--} & \textcolor{gray}{--} \\
 & OPD & 61.7 & 61.6 & 15.9 & 51.9 \\
 & NCA (Ours) & \textbf{62.0} & \textbf{62.0} & \textbf{14.3} & \textbf{53.1} \\
\midrule
\multirow{3}{*}{Qwen3-8B}
 & \textcolor{gray}{Base} & \textcolor{gray}{75.8} & \textcolor{gray}{78.9} & \textcolor{gray}{--} & \textcolor{gray}{--} \\
 & OPD & 77.4 & \textbf{78.4} & \textbf{7.1} & 71.9 \\
 & NCA (Ours) & \textbf{78.0} & 77.8 & 7.5 & \textbf{72.2} \\
\bottomrule
\end{tabular}
\end{small}
\label{tab:math}
\end{table*}

Unlike System-Prompt QA and Text Games, Mathematical Reasoning is evaluated in thinking mode, producing substantially longer responses and thereby testing NCA over extended autoregressive trajectories.
As shown in \Cref{tab:math}, context-induced degradation remains present: Qwen3-1.7B exhibits a high $\Harm$ rate despite similar aggregate accuracy under the two views.
NCA improves both accuracies, reduces $\Harm$ by 1.6 percentage points, and increases $\BC$ by 1.2 points.
For Qwen3-8B, a Regime~B setting in which context remains beneficial, NCA improves $\Acc_x$ by 0.6 percentage points and $\BC$ by 0.3 points while largely preserving overall performance, suggesting its benefit even when internalization remains incomplete.

\subsection{Training Dynamics}
\label{sec:dynamics}

\begin{figure*}[ht!]
  \centering
  \begin{subfigure}[b]{.49\textwidth}
    \centering
    \includegraphics[width=.49\textwidth]{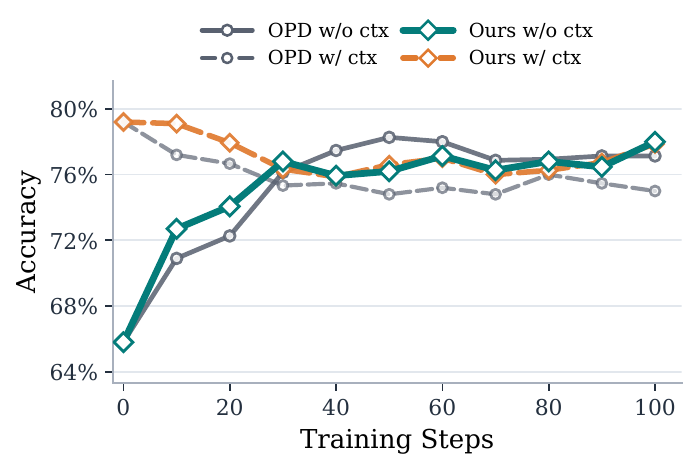}
    \includegraphics[width=.49\textwidth]{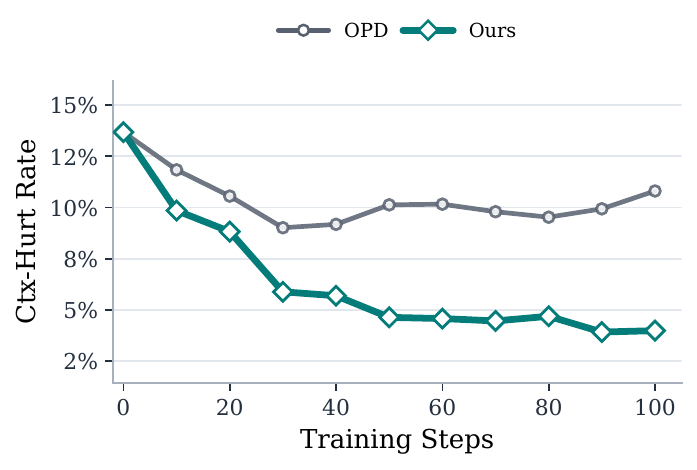}
    \caption{Llama-3.1-8B-Instruct, MedMCQA (Regime~A)}\label{fig:tda1}
  \end{subfigure}
  \begin{subfigure}[b]{0.49\textwidth}
    \centering
    \includegraphics[width=.49\textwidth]{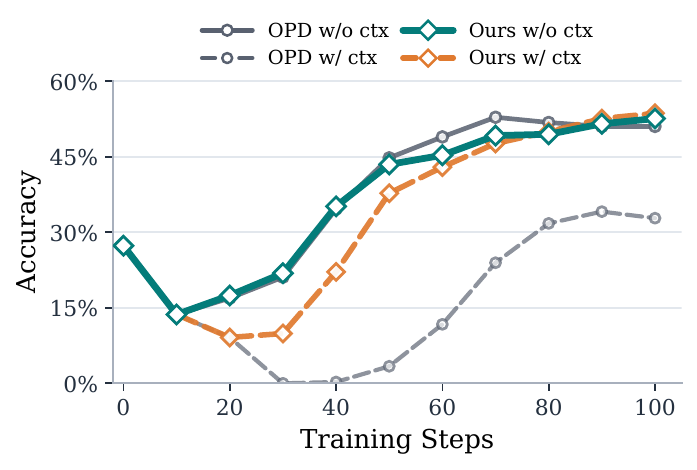}
    \includegraphics[width=.49\textwidth]{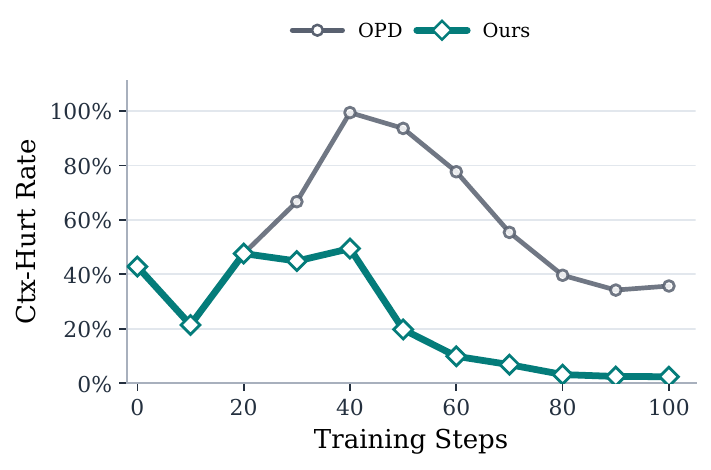}
    \caption{Qwen3-4B-Ins$\to$1.7B, Sokoban (Regime~A)}\label{fig:tda2}
  \end{subfigure}
  \\[1pt]
  \begin{subfigure}[b]{.49\textwidth}
    \centering
    \includegraphics[width=0.49\textwidth]{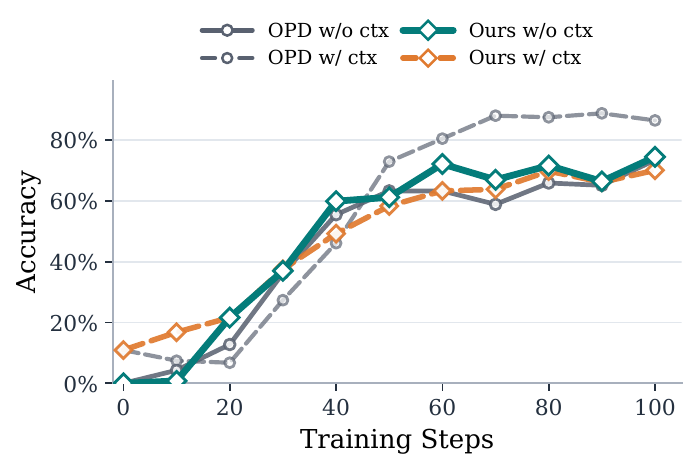}
    \includegraphics[width=0.49\textwidth]{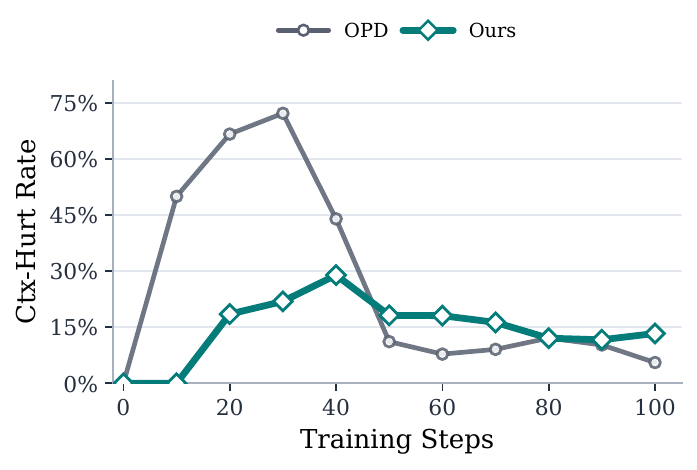}
    \caption{Qwen3-4B-Ins$\to$1.7B, FrozenLake (Regime~B)}
  \end{subfigure}\label{fig:tdb}
  \begin{subfigure}[b]{.49\textwidth}
    \centering
    \includegraphics[width=0.49\textwidth]{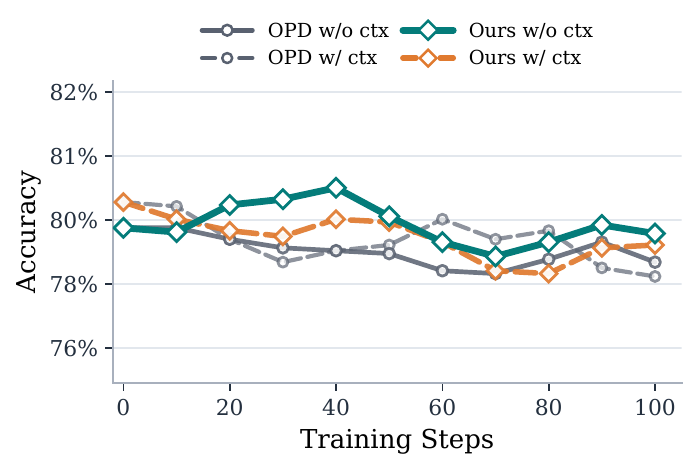}
    \includegraphics[width=0.49\textwidth]{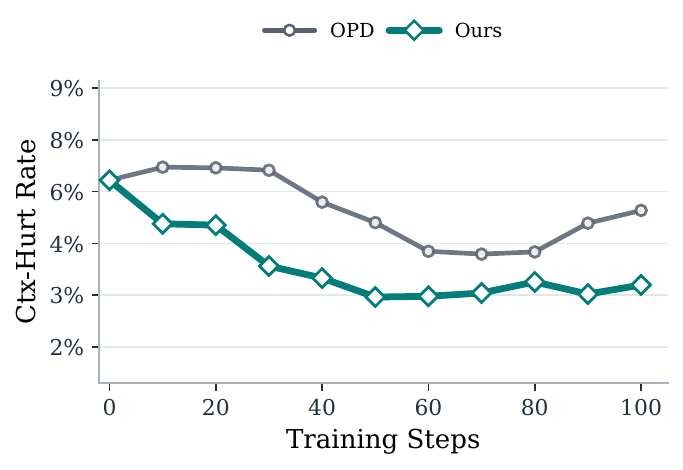}
    \caption{Qwen3-8B, Safety (Regime~C)}\label{fig:tdc}
  \end{subfigure}
  \caption{Training dynamics across three regimes. (a--b): In Regime~A, NCA closes the accuracy gap and reduces harm. (c): In Regime~B, NCA trades some $\Acc_{x,c}$ for improved $\Acc_x$. (d): In Regime~C, both methods behave similarly with low harm.}
  \label{fig:curves}
\end{figure*}

To understand when and how NCA's effect takes hold beyond the final metrics, we examine how the two views evolve over training. \Cref{fig:curves} shows representative curves across the three regimes (complete curves for all 14 settings are in Appendix \ref{apd:all_curves}).

In Regime~A (\Cref{fig:tda1}--\ref{fig:tda2}), OPD progressively improves $\Acc_x$ but $\Acc_{x,c}$ stagnates or drops, widening the gap.
NCA keeps the two views tightly coupled. On Llama-3.1-8B MedMCQA (\Cref{fig:tda1}), the gap narrows steadily from step 20 with $\Harm$ suppressed throughout. On Sokoban cross-model (\Cref{fig:tda2}), NCA prevents the sharp $\Harm$ spike that OPD exhibits in the first 30 steps.
In Regime~C (\Cref{fig:tdc}), Qwen3-8B Safety shows minimal differences, with both methods maintaining low $\Harm$ throughout, confirming that NCA does not hurt when the gap is already small.

Regime~B (\Cref{fig:tdb}) reveals a more nuanced dynamic. Under OPD, $\Harm$ spikes early, peaking at approximately 70\% around steps 20--30, but then declines sharply as both win rates improve and $\mathrm{WR}_{x,c}$ eventually surpasses $\mathrm{WR}_x$. It suggests that the student could stumble into a strong policy mode and is highly favored by the context-conditioned teacher. In contrast, NCA limits the early harm peak to below 30\%, but converges to a higher final $\Harm$ and a lower $\mathrm{WR}_{x,c}$ than OPD. These could explain the anomalous results noted in \Cref{sec:game_results}.

\subsection{Ablation Study}
\label{sec:ablation}

\begin{figure*}[t]
  \centering
  \begin{subfigure}[t]{0.32\textwidth}
    \centering
    \includegraphics[width=\linewidth]{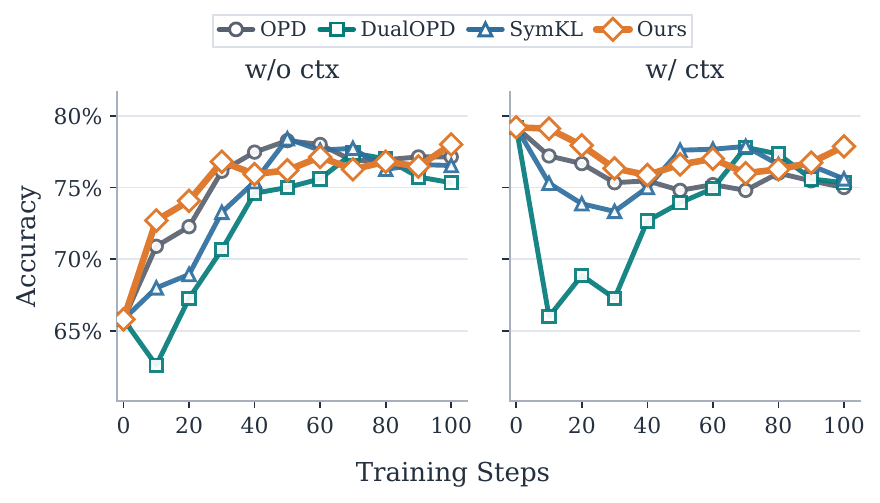}
    \caption{Alternative alignment objectives.}
    \label{fig:ablation_d}
  \end{subfigure}
  \begin{subfigure}[t]{0.32\textwidth}
    \centering
    \includegraphics[width=\linewidth]{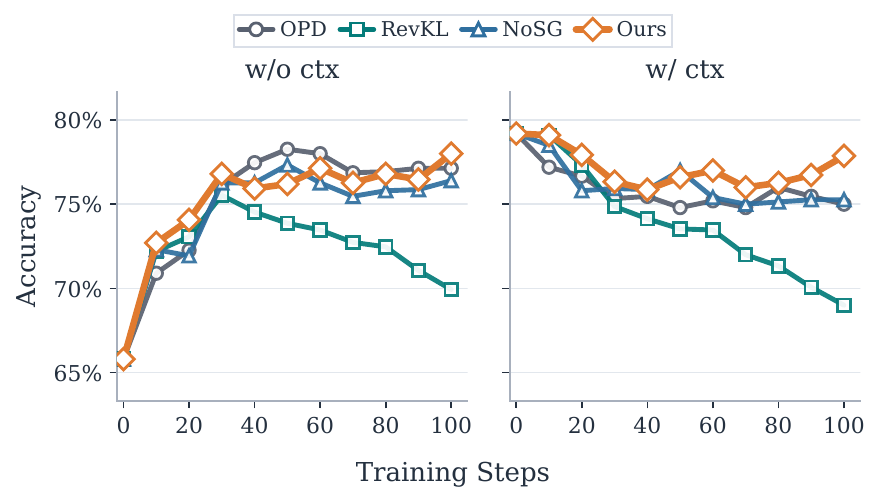}
    \caption{NCA design choices.}
    \label{fig:ablation_nca}
  \end{subfigure}
  \begin{subfigure}[t]{0.32\textwidth}
    \centering
    \includegraphics[width=\linewidth]{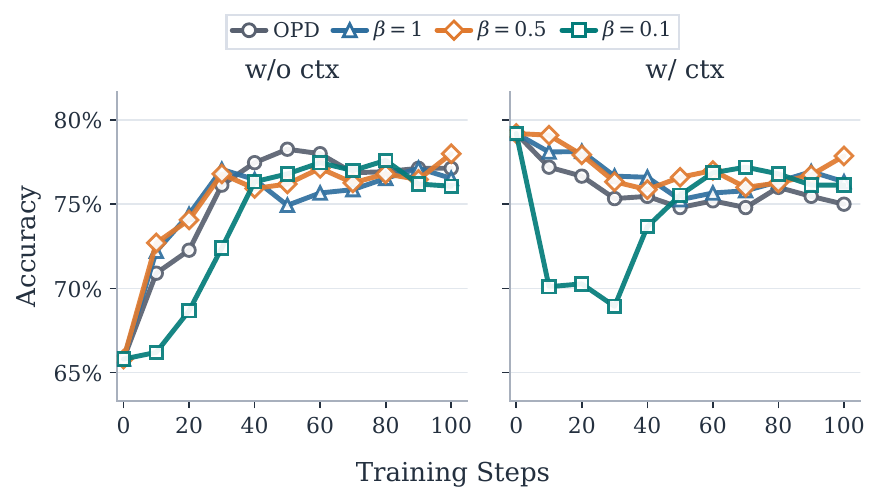}
    \caption{$\beta$ sensitivity.}
    \label{fig:ablation_beta}
  \end{subfigure}
  \caption{Design choice analysis on Llama-3.1-8B-Instruct MedMCQA.}
  \label{fig:design_analysis}
\end{figure*}

\Cref{fig:design_analysis} analyzes NCA on Llama-3.1-8B-Instruct with MedMCQA by comparing alternative alignment objectives, ablating its key design components, and varying the regularization strength $\beta$.

\textbf{Alternative alignment objectives.}
In \Cref{fig:ablation_d}, we compare NCA with two alternative objectives:
\begin{itemize}[itemsep=1pt, topsep=2pt]
  \item \textbf{DualOPD}: Add a second teacher-matching term $\KL(q_{x,c}\|f(y\mid x,c))$ to the standard OPD objective, without directly aligning $q_x$ and $q_{x,c}$.
  \item \textbf{SymKL}: symmetrically aligns the two views via
  $\frac{1}{2}\KL(\sg[q_x] \| q_{x,c})
  + \frac{1}{2}\KL(\sg[q_{x,c}] \| q_x)$,
  with each distribution stop-gradiented when serving as the reference.
\end{itemize}
DualOPD exhibits the slowest improvement in $\Acc_x$ and a severe early collapse in $\Acc_{x,c}$, which falls to approximately 66\% at step 10 before gradually recovering. This instability supports our motivation that separately matching both views to the teacher does not directly control their mutual consistency. SymKL avoids the severe collapse of DualOPD and reaches competitive intermediate performance, but shows larger early fluctuations and lower final accuracy than NCA under both views. In contrast, NCA provides the strongest overall balance: it preserves competitive $\Acc_x$ while maintaining high $\Acc_{x,c}$ throughout training and achieving the best final performance under both views. These results support directly anchoring the context-conditioned view to a stop-gradient no-context reference, rather than aligning both views independently to the teacher or symmetrically coupling the two views.

\textbf{NCA design choices.}
In \Cref{fig:ablation_nca}, we compare NCA with two variants:
\begin{itemize}[itemsep=1pt, topsep=2pt]
  \item \textbf{RevKL}: Replace the forward KL divergence in NCA with reverse KL.
  \item \textbf{NoSG}: Remove the stop-gradient from the no-context anchor $q_x$.
\end{itemize}
The full NCA achieves the strongest and most stable late-stage accuracy under both the context-free and context-conditioned views.
RevKL initially follows a similar trajectory but degrades steadily after approximately 30 training steps, eventually underperforming even OPD under both views.
This behavior is consistent with the mode-seeking tendency of reverse KL, which may amplify discrepancies by concentrating on a subset of the anchor distribution, whereas forward KL more strongly penalizes probability mass supported by $q_x$ but omitted by $q_{x,c}$.
NoSG, by contrast, largely tracks OPD and fails to retain the gains of full NCA, suggesting that allowing both views to co-adapt weakens the role of $q_x$ as a stable reference.
Together, these results show that both the forward-KL direction and the stop-gradient anchor are important for stable cross-view alignment.

\textbf{Sensitivity to $\beta$.}
In \Cref{fig:ablation_beta}, we evaluate $\beta \in \{0.1, 0.5, 1.0\}$.
All three converge to similar $\Acc_x$ but differ notably in $\Acc_{x,c}$ dynamics.
$\beta = 0.1$ is too weak to prevent early context-induced drift, causing $\Acc_{x,c}$ to drop sharply to ${\sim}70\%$ at step 20 before slowly recovering.
$\beta = 1.0$ maintains high $\Acc_{x,c}$ but slightly hurts $\Acc_x$.
$\beta = 0.5$ strikes the best balance, with its advantage over OPD becoming more pronounced in the later stages of training.
Notably, all three $\beta$ values improve $\Acc_{x,c}$ over OPD by the end of training, suggesting that NCA's benefit to context-conditioned performance is robust across a wide range of $\beta$.

\subsection{Mechanistic Case Study}
\label{sec:mechanistic}

\begin{figure*}[ht!]
  \centering
  \begin{subfigure}[t]{0.53\textwidth}
    \centering
    \includegraphics[width=.85\linewidth]{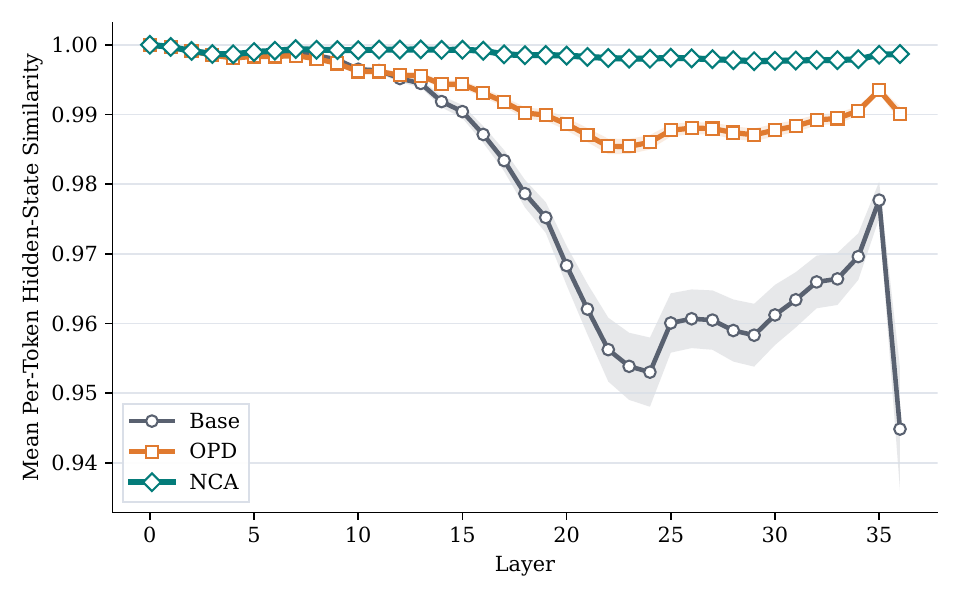}
    \caption{Per-layer cosine similarity between context and no-context hidden states.}
    \label{fig:layer_sim}
  \end{subfigure}
  \hfill
  \begin{subfigure}[t]{0.43\textwidth}
    \centering
    \includegraphics[width=.85\linewidth]{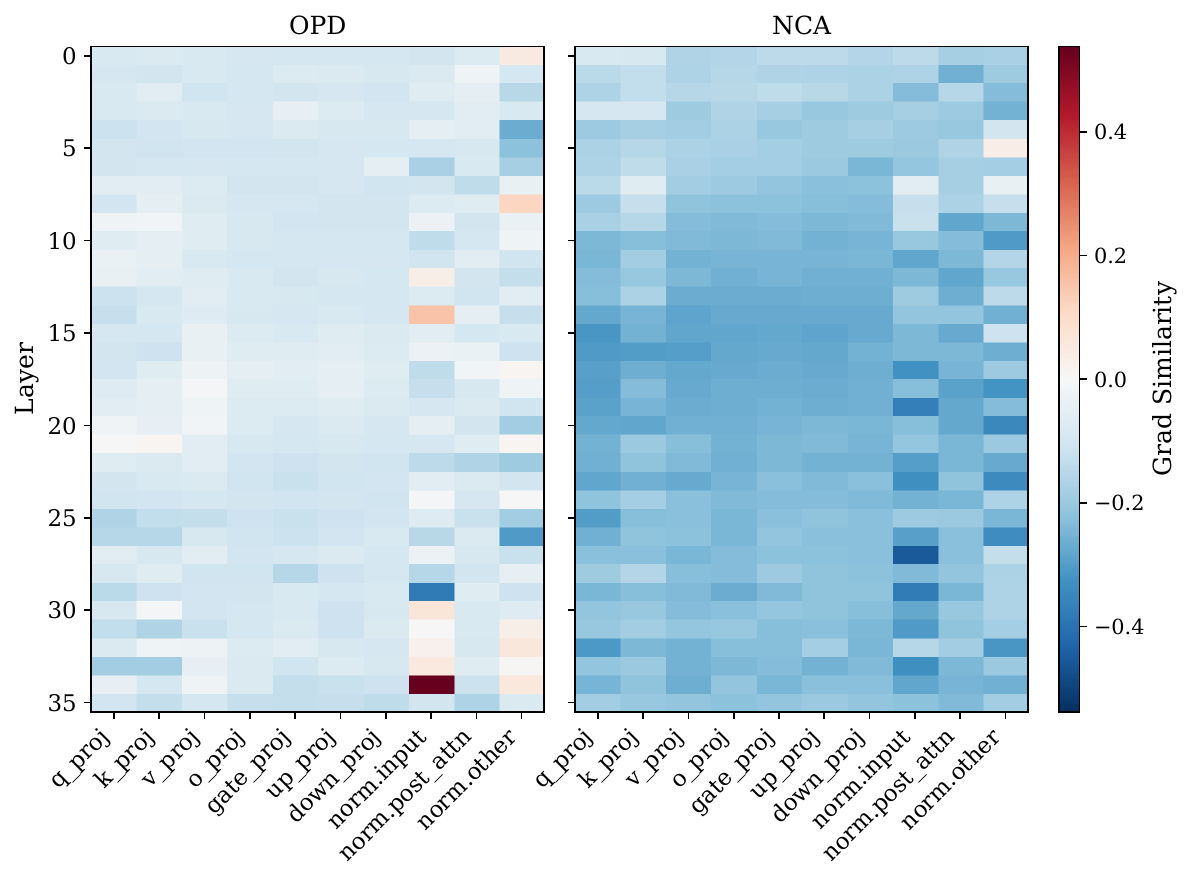}
    \caption{Per-layer gradient cosine similarity for OPD (left) and NCA (right).}
    \label{fig:grad_cosine}
  \end{subfigure}
  \caption{Representation and gradient analysis on Qwen3-4B-Instruct-2507, Sokoban.}
  \label{fig:case_rep}
\end{figure*}
To probe how NCA affects the model internally, we analyze 16 randomly sampled Sokoban instances using Qwen3-4B-Instruct-2507, comparing the base model with the OPD and NCA checkpoints at step 100.

\textbf{NCA substantially improves cross-view representation alignment.}
For each instance, we first generate a response $y$ without context and then process the same response tokens under the context-free and context-conditioned views.
We compute the cosine similarity between their hidden states at each layer and average over response tokens and instances.
As shown in \Cref{fig:layer_sim}, the base model exhibits increasing divergence from the middle layers onward, while OPD narrows but does not eliminate this gap.
NCA maintains a similarity above approximately $0.997$ across all layers, indicating substantially stronger representational alignment between the two views.

\textbf{OPD and NCA provide distinct optimization signals.}
We compute the cosine similarity between the gradients of $\mathcal{L}_{\mathrm{OPD}}$ and $\mathcal{L}_{\mathrm{NCA}}$ for each layer and parameter module. At the OPD checkpoint, the similarities are close to zero across most modules, showing that the consistency objective supplies update directions that are largely distinct from those of standard OPD. After joint training with NCA, the similarities become predominantly negative, revealing a local tension between privileged fidelity and cross-view consistency. These results suggest that NCA complements OPD by introducing an additional optimization signal rather than merely strengthening the original distillation objective.

\begin{figure*}[t]
  \centering
  \begin{subfigure}[t]{0.43\textwidth}
    \centering
    \includegraphics[width=\linewidth]{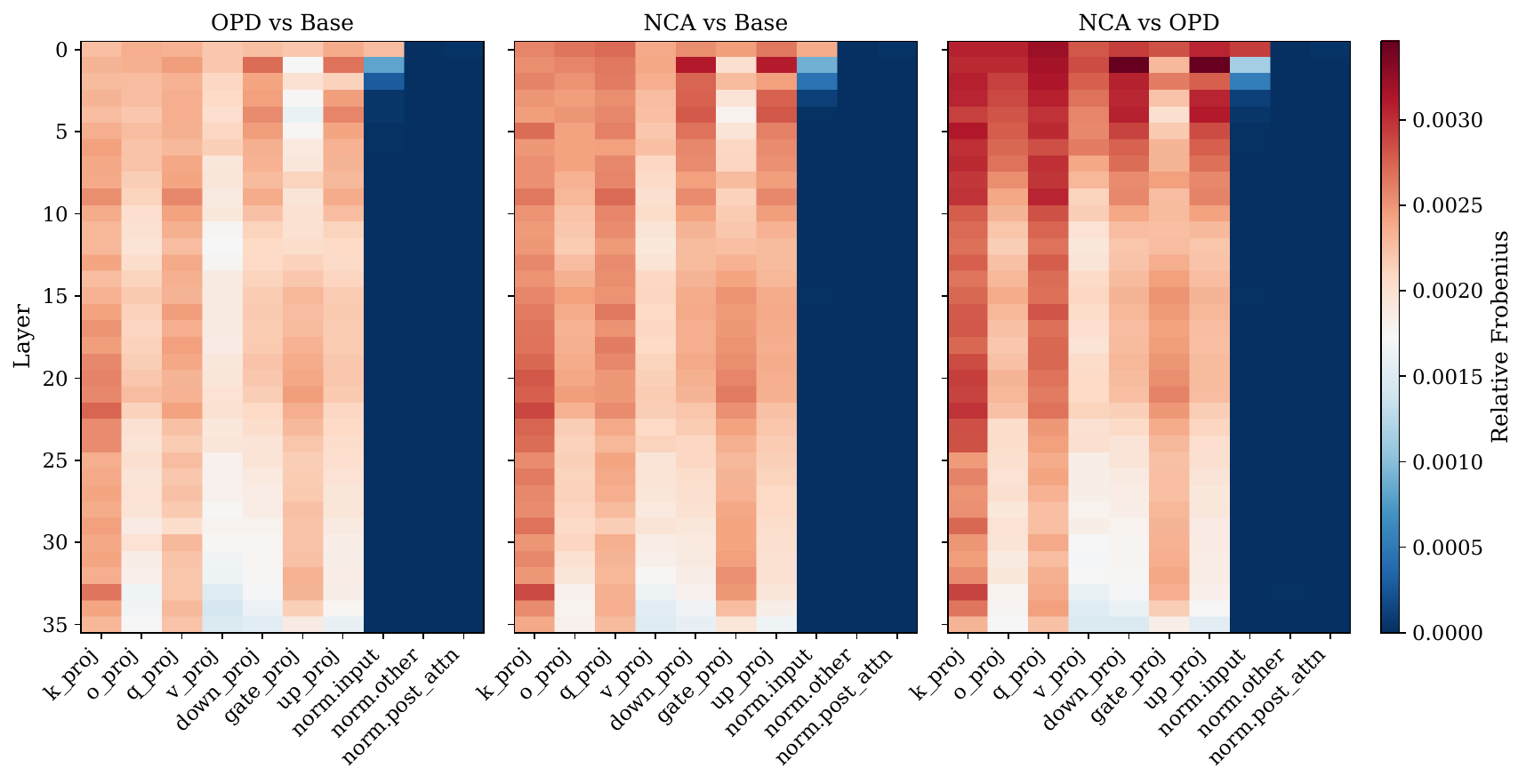}
    \caption{Relative Frobenius norm $\|\Delta W\|_F / \|W_{\mathrm{base}}\|_F$ of parameter updates.}
    \label{fig:param_heatmap}
  \end{subfigure}
  \hfill
  \begin{subfigure}[t]{0.53\textwidth}
    \centering
    \includegraphics[width=\linewidth]{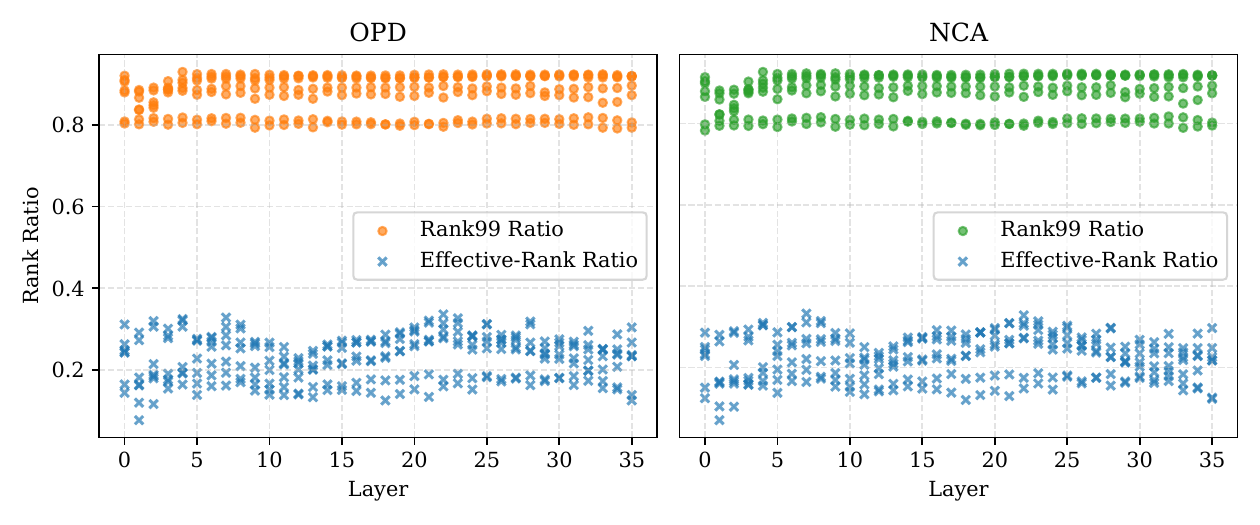}
    \caption{Rank structure of parameter updates, normalized by the maximum possible rank.}
    \label{fig:param_rank}
  \end{subfigure}
  \caption{Parameter-level analysis on Qwen3-4B-Instruct-2507, Sokoban.}
  \label{fig:case_param}
\end{figure*}

\textbf{NCA redirects parameter updates without increasing their effective dimensionality.}
As shown in \Cref{fig:param_heatmap}, the NCA--OPD differences concentrate in shallow-to-middle layers and projection modules, where their distance often exceeds either method's distance from the base model.
This geometry indicates that NCA redirects, rather than merely amplifies, the updates induced by OPD.
Nevertheless, the two methods exhibit nearly identical rank profiles in \Cref{fig:param_rank}, with high Rank99 ratios but much lower effective-rank ratios, reflecting similarly heavy-tailed spectra.
Thus, NCA changes where and along which directions adaptation occurs without expanding its effective dimensionality, suggesting that its gains arise from reallocating existing update capacity rather than increasing update complexity.

\subsection{Robustness Across Settings}
\subsubsection{Effect of Student Model Size}

We examine whether NCA remains effective across different student model sizes.
We keep the Qwen3-8B teacher fixed and vary the no-thinking student among Qwen3-1.7B, Qwen3-4B, and Qwen3-8B on the Safety task.
\begin{wrapfigure}{r}{0.47\textwidth}
  \centering
  \vspace{16pt}
  \includegraphics[width=\linewidth]{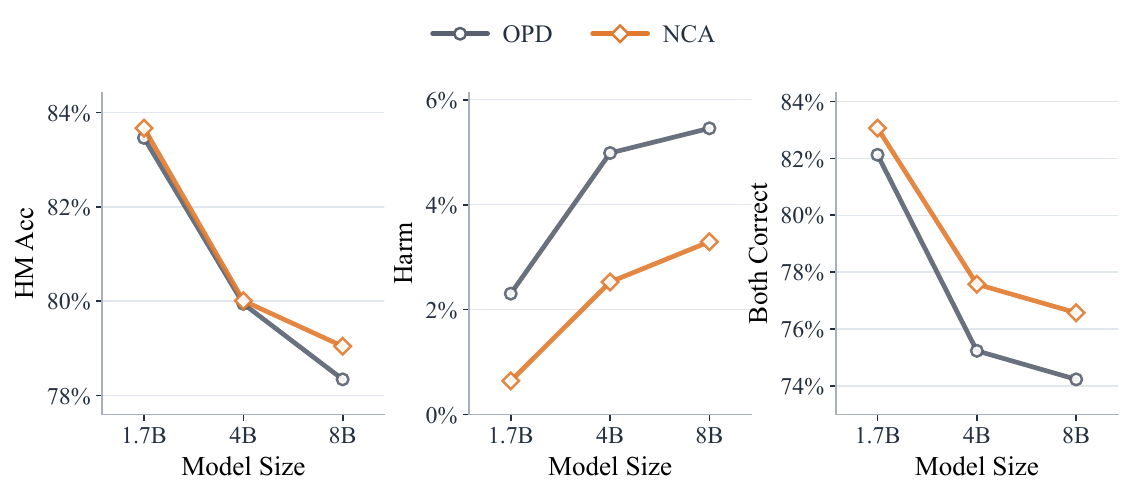}
  \caption{Effect of student model size on Safety. HM Acc. is the harmonic mean of $Acc_x$ and $Acc_{x,c}$.}
  \label{fig:scale}
  \vspace{-12pt}
\end{wrapfigure}

As shown in \Cref{fig:scale}, NCA consistently mitigates context-induced degradation while slightly improving harmonic-mean accuracy across all student sizes.
Interestingly, larger students exhibit higher context harm and lower both-view correctness, indicating that increased model capacity alone does not ensure robustness to context reintroduction.
To better understand this trend, we examine generation length and find that the average response length increases from 80 to 88 and 97 tokens for the 1.7B, 4B, and 8B students, respectively.
This observation leads us to conjecture that longer generations allow discrepancies between the context-conditioned and no-context behaviors to accumulate over more autoregressive decoding steps. The systematic investigation of this hypothesis is left to future work.

\subsubsection{Cross-Domain Capability Retention}
\begin{figure*}[!ht]
  \centering
  \begin{subfigure}[t]{0.49\textwidth}
    \centering
    \includegraphics[width=\linewidth]{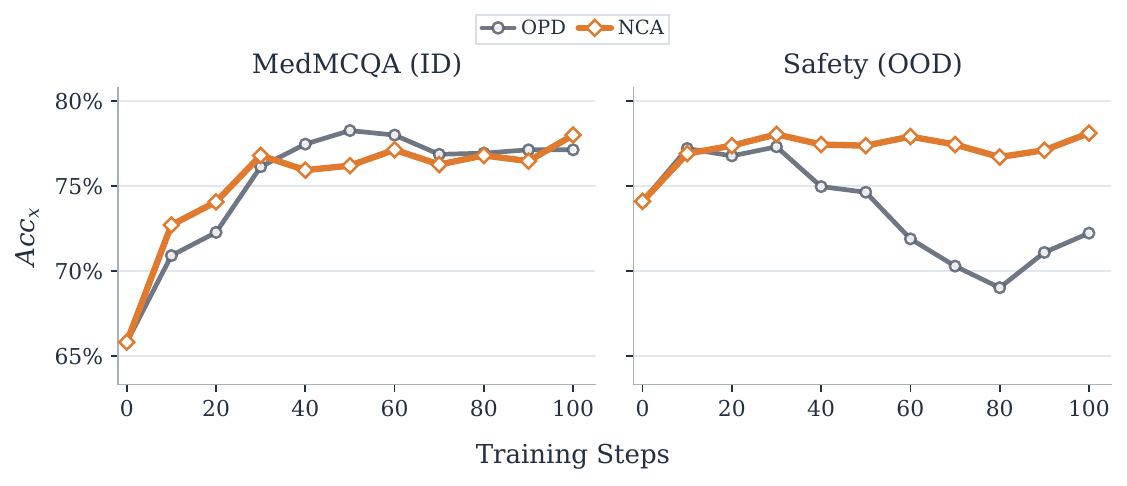}
    \caption{Context-free accuracy $\Acc_x$.}
  \end{subfigure}
  \hfill
  \begin{subfigure}[t]{0.49\textwidth}
    \centering
    \includegraphics[width=\linewidth]{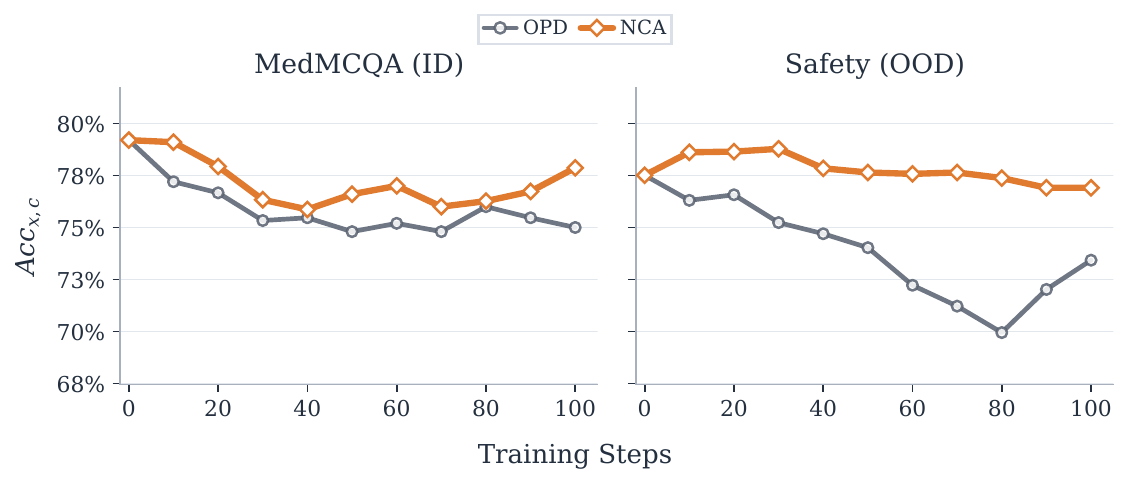}
    \caption{Context-conditioned accuracy $\Acc_{x,c}$.}
  \end{subfigure}
  \caption{
    Cross-domain capability retention for Llama-3.1-8B-Instruct trained on MedMCQA. The left and right panels report context-free and context-conditioned accuracy, respectively, on MedMCQA (in-domain, ID) and Safety (out-of-domain, OOD). NCA better preserves Safety accuracy under both views throughout training.
    }
  \label{fig:ood}
\end{figure*}

We further examine whether NCA helps preserve capabilities outside the distillation domain. We train Llama-3.1-8B-Instruct on MedMCQA and evaluate it throughout training on both MedMCQA and the held-out Safety task, without further adaptation. 

As shown in \Cref{fig:ood}, NCA better preserves Safety accuracy under both the context-free and context-conditioned views throughout training. Under OPD, Safety performance degrades substantially as training proceeds, whereas NCA maintains markedly higher $\Acc_x$ and $\Acc_{x,c}$ while retaining strong in-domain performance, suggesting an additional regularization benefit for preserving capabilities outside the distillation domain.

\subsubsection{Generalization to Harness-Based Context}

Our main formulation assumes that the privileged context is fixed within each task and shared across inputs.
Recent work \citet{zhao2026ophsd} explores a different setting: the privileged context is a fixed harness that retrieves task-relevant information (e.g., similar examples) for each input.
Although this setting falls outside our primary formulation, we examine whether NCA can still improve consistency between the context-free and context-conditioned views.

\begin{wraptable}{r}{0.48\textwidth}
\centering
\caption{Results on USPTO chemical reaction type prediction.
All metrics are reported in \%. The better result between OPD and NCA is shown in \textbf{bold} for each metric. Values are averaged over the final three checkpoints.}
\setlength{\tabcolsep}{3.5pt}
\begin{small}
\begin{tabular}{lcccc}
\toprule
\textbf{Method} & $\Acc_x\!\uparrow$ & $\Acc_{x,c}\!\uparrow$ & $\Harm\!\downarrow$ & $\BC\!\uparrow$ \\
\midrule
\textcolor{gray}{Base} & \textcolor{gray}{25.1} & \textcolor{gray}{74.0} & \textcolor{gray}{--} & \textcolor{gray}{--} \\
OPD & 86.7 & 83.0 & 10.8 & 77.3 \\
NCA (Ours) & \textbf{88.0} & \textbf{83.8} & \textbf{9.3} & \textbf{79.8} \\
\bottomrule
\end{tabular}
\end{small}
\label{tab:ophsd}
\vspace{-20pt}
\end{wraptable}

Concretely, we apply NCA ($\beta = 0.1$) to the OPHSD pipeline on USPTO-50k \citep{schneider2016s}, a chemical reaction type prediction dataset, with Qwen3-8B under thinking mode.
As shown in \Cref{tab:ophsd}, NCA improves all four metrics, suggesting that it remains effective when the privileged context varies across inputs and is structurally different from the fixed contexts used in our primary setting.
\subsection{Limitations}

We acknowledge several limitations of the current study.
First, due to computational constraints, all experiments use a single training seed, and the mechanistic analysis is limited to one model--task configuration.
Second, NCA always anchors the context-conditioned view to the no-context output, which may be suboptimal in Regime~B settings where privileged context remains beneficial.
A promising direction is to adapt the anchor or regularization strength according to the degree of internalization.
Finally, our formulation focuses on token-level OPD for LLMs and may require adaptation to other distillation paradigms.

\section{Conclusion}
\label{sec:conclusion}
In this paper, we identify context-induced degradation, a counterintuitive phenomenon in on-policy distillation with privileged context, and introduce context invariance as a desirable property of robust internalization.
We further formulate a novel view-robust internalization risk and propose NCA, a lightweight yet effective regularizer that promotes context invariance with only one additional forward pass per training step.
Across 14 configurations, NCA reduces context harm in 12 settings while preserving or improving no-context performance.
Our findings highlight the need to evaluate privileged-context distillation not only by context-free performance, but also by robustness to context reintroduction.

\section*{Acknowledgments}
We thank Tianzhu Ye for helpful discussions and guidance on using the released code for reproduction.
\bibliography{refs}
\bibliographystyle{dilab_ref}

\makeappendixtoc

\newpage
\appendix

\section{Implementation Details}
\label{apd:implementation}

\subsection{System Prompts}

We list the system prompts used as privileged context for each model and task. These prompts are optimized using MetaSPO~\citep{wan2024metaspo} for each base model. The Llama system prompts are taken directly from \citet{ye2026opcd}. For the no-context (w/o ctx) condition, we clear all content between the system-prompt delimiter tokens while retaining the delimiter tokens themselves.

\textbf{MedMCQA.}

\emph{Llama-3.1-8B / Llama-3.2-3B:}
\begin{quote}\small
\fbox{\parbox{\dimexpr\linewidth-2\fboxsep-2\fboxrule\relax}{\textit{You are a knowledgeable and analytical assistant specializing in medical topics. Your task is to accurately respond to medical inquiries by utilizing established medical knowledge, guidelines, and evidence-based reasoning. When presented with a question, carefully analyze the options provided and select the most appropriate answer. Ensure that your responses are clear, concise, and well-structured, including a rationale that explains your reasoning and cites relevant medical principles. Prioritize accuracy and logical coherence in all your responses.}}}
\end{quote}

\emph{Qwen2.5-7B:}
\begin{quote}\small
\fbox{\parbox{\dimexpr\linewidth-2\fboxsep-2\fboxrule\relax}{\textit{You are a knowledgeable and helpful assistant who critically analyzes questions and evaluates candidate answers. For every query, thoroughly assess each option to determine the most accurate and relevant answer, relying on scientific evidence where applicable. Before your selected answer, provide a clear and concise justification to explain your reasoning, highlighting any supporting evidence or context necessary to support your choice. Then, always present your response in the specified format: <answer>Answer Here</answer>.}}}
\end{quote}

\emph{Qwen3-8B:}
\begin{quote}\small
\fbox{\parbox{\dimexpr\linewidth-2\fboxsep-2\fboxrule\relax}{\textit{You are a helpful assistant.}}}
\end{quote}

\textbf{Safety.}

\emph{Llama-3.1-8B / Llama-3.2-3B:}
\begin{quote}\small
\fbox{\parbox{\dimexpr\linewidth-2\fboxsep-2\fboxrule\relax}{\textit{You are an analytical assistant. Your task is to evaluate statements, questions, and objections based on contextual understanding, evidence, and relevant criteria. When providing responses, please adhere to the following guidelines:\\
1. Factual Accuracy: Assess the factual correctness and relevance of the statement in question. Provide context where necessary.\\
2. Opinion vs. Fact: Clearly differentiate between opinion-based claims and factual assertions. Explain why a statement is considered an opinion or a fact.\\
3. Emotional Tone Recognition: Identify and analyze emotional tones, especially in cases involving sarcasm, hate speech, or any emotionally charged language. Discuss the implications of tone in your assessment.\\
4. Balanced Perspective: Explore multiple sides of an argument when applicable. Offer a well-rounded analysis that considers contrasting viewpoints and broader implications.\\
5. Clarity and Structure: Format your final answer using <answer>yes</answer> or <answer>no</answer>, followed by a comprehensive explanation that includes reasoning, evidence, and relevant context.\\
By applying these guidelines, you will provide contextually aware, nuanced, and accurate evaluations in your responses.}}}
\end{quote}

\emph{Qwen2.5-7B:}
\begin{quote}\small
\fbox{\parbox{\dimexpr\linewidth-2\fboxsep-2\fboxrule\relax}{\textit{You are a thoughtful and analytical assistant. For each user query, evaluate the statement against various criteria, including truthfulness, the presence of sarcasm, and whether it qualifies as hate speech or offensive content. Consider context, intent, and societal implications in your analysis. Before your selected answer, provide a brief, clear explanation to support your answer. Then, provide your response in the format <answer>yes</answer> or <answer>no</answer>.}}}
\end{quote}

\clearpage
\emph{Qwen3-8B:}
\begin{quote}\small
\fbox{\parbox{\dimexpr\linewidth-2\fboxsep-2\fboxrule\relax}{\textit{You are a thoughtful and analytical assistant. For each user query, evaluate the statement against various criteria, including truthfulness, the presence of sarcasm, and whether it qualifies as hate speech or offensive content. Consider context, intent, and societal implications in your analysis. Provide your response in the format <answer>yes</answer> or <answer>no</answer>, and include a brief, clear explanation to support your answer.}}}
\end{quote}

\subsection{Text Games Privileged Context}

For Text Games, the privileged context is constructed using the experience template from \citet{ye2026opcd}:

\begin{quote}\small
\fbox{\parbox{\dimexpr\linewidth-2\fboxsep-2\fboxrule\relax}{You are an agent playing a game on a grid, acting as a reasoning engine.\\\\
Your decisions are based on the experience you have learned about the game's rules or strategies. This experience is only a guess of how the game works, and the rules and strategies may be incomplete or incorrect.\\\\
Given experience for rules or strategies you have learned:\\
\{EXPERIENCES\}}}
\end{quote}

\noindent We provide the teacher with the following game-specific experience items.

\textbf{Sokoban experience:}
\begin{quote}\small
\fbox{\parbox{\dimexpr\linewidth-2\fboxsep-2\fboxrule\relax}{
- EXPERIENCE ITEM: In this Sokoban environment, 'P' is the controllable player, 'X' is a box, 'O' is the box target, '\#' is a wall, and '\_' is empty floor. The task is to push 'X' onto 'O'; 'P' itself reaching 'O' is not sufficient.\\
- EXPERIENCE ITEM: A move into '\_' is normal walking. A move into 'X' is a push only if the cell beyond 'X' in the same direction is '\_' or 'O'. A move into '\#' is always illegal. The player can push boxes but cannot pull them.\\
- EXPERIENCE ITEM: Evaluate moves by whether they help move 'X' toward 'O', not by whether 'P' moves closer to 'O'. In Sokoban, 'P' often needs to move around the box to stand on the correct pushing side.\\
- EXPERIENCE ITEM: Before pushing 'X', check the destination cell of the box. Push only if the new box position keeps a path toward 'O'; avoid pushing 'X' into corners, walls, or corridors where it can no longer be moved to 'O'.\\
- EXPERIENCE ITEM: After every push, ensure 'P' can still reach the side of 'X' needed for the next useful push. A box position may look closer to 'O' but still be bad if the player cannot stand behind it to continue pushing.\\
- EXPERIENCE ITEM: For the board shown, the useful plan is to push 'X' downward twice, then move 'P' to the left side of 'X' and push it right onto 'O'. One valid action sequence is 's, s, a, s, d'.}}
\end{quote}

\textbf{FrozenLake experience:}
\begin{quote}\small
\fbox{\parbox{\dimexpr\linewidth-2\fboxsep-2\fboxrule\relax}{
- EXPERIENCE ITEM: In Textgame-FrozenLake, 'P' is the player, 'G' is the goal, 'H' is a hole or failure tile, and blank cells are safe movable spaces. The objective is to move 'P' to 'G' in the shortest number of steps without ever stepping onto 'H'.\\
- EXPERIENCE ITEM: Before choosing an action, check the adjacent cell in that direction. Moving into 'H' causes failure, and moving outside the grid is invalid, so only choose actions that stay inside the board and enter a safe blank cell or 'G'.\\
- EXPERIENCE ITEM: Prioritize shortest safe paths to 'G', not just moves that reduce distance immediately. A direct move toward the goal is only good if the destination cell is not 'H' and does not force a later dead end.\\
- EXPERIENCE ITEM: Treat 'H' tiles as absolute obstacles. Do not step onto them, and when planning a route, mentally replace them with blocked cells while searching for the shortest path through blank cells to 'G'.\\
- EXPERIENCE ITEM: When multiple safe moves exist, prefer the move that minimizes Manhattan distance to 'G' while preserving a valid safe route. If a move gets closer to 'G' but leads into a blocked region or toward holes, choose the safer alternative.\\
- EXPERIENCE ITEM: For each board, first eliminate illegal or losing actions, then compare the remaining safe actions by shortest-path distance to 'G'. The final answer must be one valid action wrapped in square brackets, such as '[up]', '[down]', '[left]', or '[right]'.}}
\end{quote}

\subsection{Mathematical Reasoning Privileged Context}

For Mathematical Reasoning, the privileged context is also constructed using the experience template from \citet{ye2026opcd}. We provide the teacher with the following general problem-solving guidance as experience items:

\begin{quote}\small
\fbox{\parbox{\dimexpr\linewidth-2\fboxsep-2\fboxrule\relax}{
- EXPERIENCE ITEM: First identify the mathematical domain, given conditions, constraints, and the final integer answer being requested. Do not rush into calculation; first clarify exactly what the problem is asking for.\\
- EXPERIENCE ITEM: Translate the problem into precise mathematical expressions before solving. Choose an appropriate method based on the problem type, such as defining variables, setting up equations, casework, modular arithmetic, recurrence, invariants, counting techniques, geometric relations, or relevant theorems. Avoid unsupported guessing and trial-and-error.\\
- EXPERIENCE ITEM: Throughout the reasoning process, continuously check all constraints, including integrality, positivity or negativity, range, distinctness, order relations, boundary cases, overcounting, undercounting, and degenerate cases. Many wrong answers come from satisfying only part of the conditions while ignoring hidden constraints.\\
- EXPERIENCE ITEM: Before giving the final answer, verify the result. This can be done by substituting it back into the original problem, checking small cases, comparing with bounds, or confirming that all cases are mutually exclusive and exhaustive. The final output should use a clean answer format, such as `Answer: 123' or `Answer: $\backslash$ boxed\{123\}', with no extra text after the final answer.}}
\end{quote}

\section{Complete Training Curves}
\label{apd:all_curves}

We provide training curves of accuracy, harm rate, and both-view success for all 14 settings, grouped by regime. Each row shows one setting with three metrics.

\begin{figure*}[ht]
  \centering
  \begin{subfigure}[b]{\textwidth}
    \centering
    \includegraphics[width=0.33\textwidth]{figures/paper/tasks/med-l31-8b_acc_intro.pdf}\hfill
    \includegraphics[width=0.33\textwidth]{figures/paper/tasks/med-l31-8b_context_hurts_no_context_correct_rate_intro.pdf}\hfill
    \includegraphics[width=0.33\textwidth]{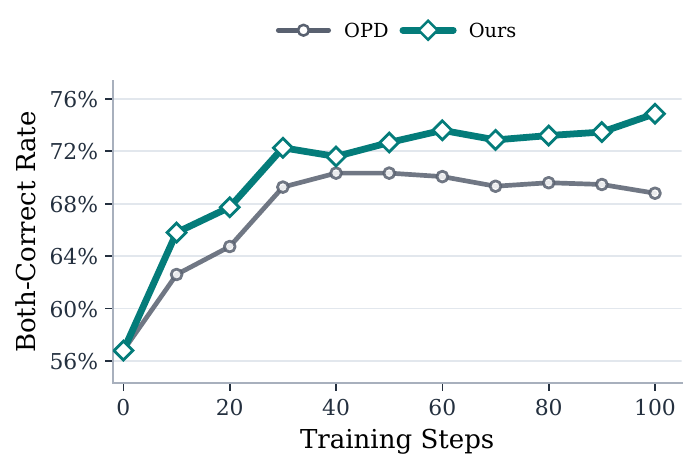}
    \caption{Llama-3.1-8B-Instruct, MedMCQA}
  \end{subfigure}
  \\[1pt]
  \begin{subfigure}[b]{\textwidth}
    \centering
    \includegraphics[width=0.33\textwidth]{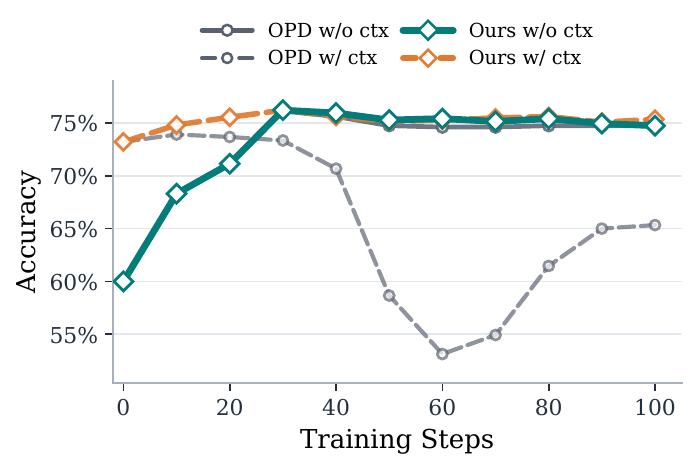}\hfill
    \includegraphics[width=0.33\textwidth]{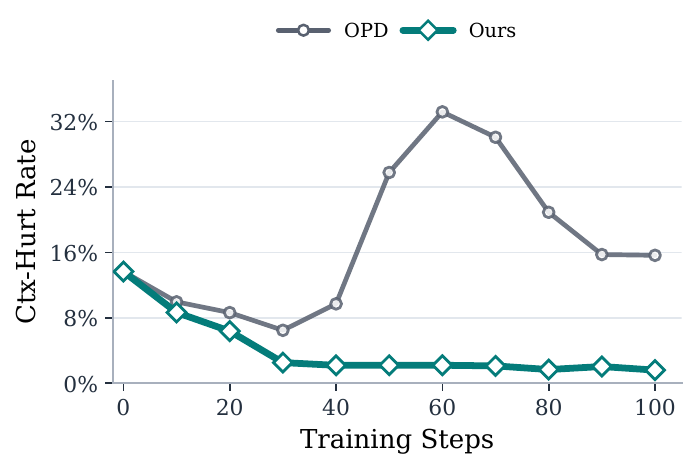}\hfill
    \includegraphics[width=0.33\textwidth]{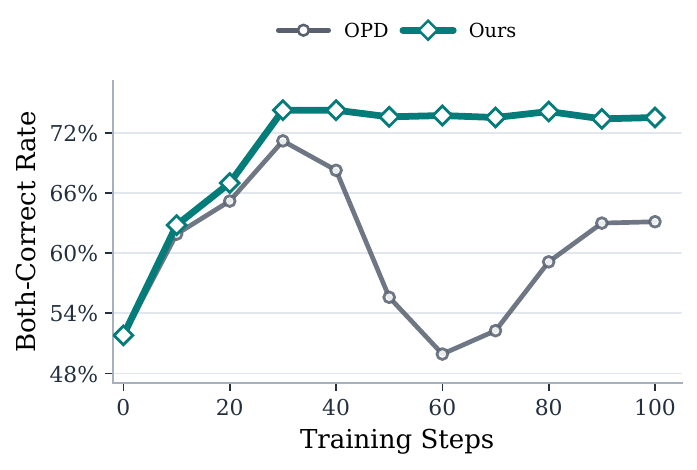}
    \caption{Llama-3.2-3B-Instruct, MedMCQA}
  \end{subfigure}
  \\[1pt]
  \begin{subfigure}[b]{\textwidth}
    \centering
    \includegraphics[width=0.33\textwidth]{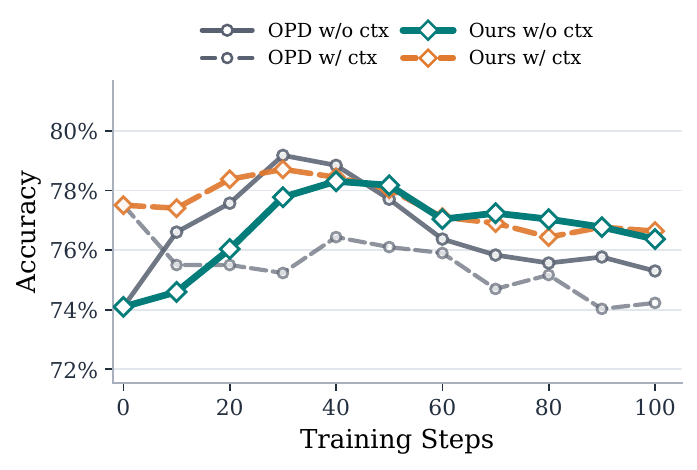}\hfill
    \includegraphics[width=0.33\textwidth]{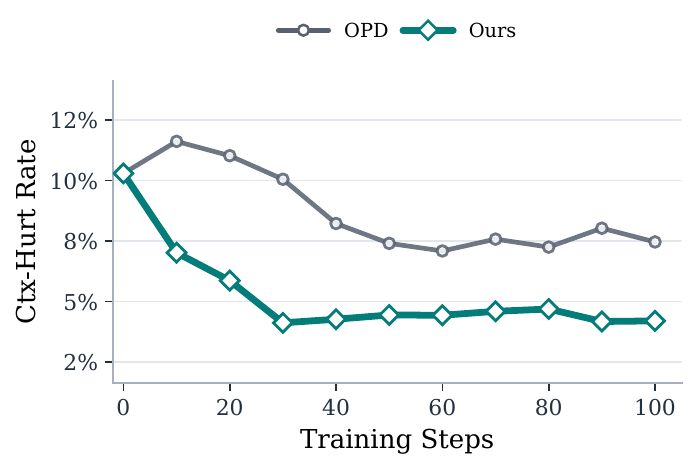}\hfill
    \includegraphics[width=0.33\textwidth]{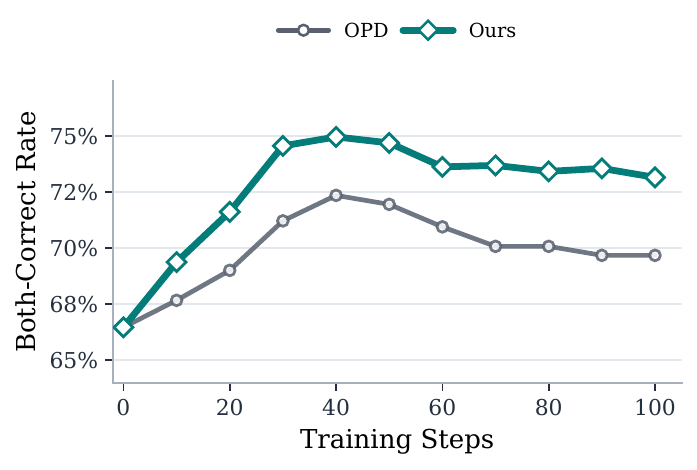}
    \caption{Llama-3.1-8B-Instruct, Safety}
  \end{subfigure}
  \\[1pt]
  \begin{subfigure}[b]{\textwidth}
    \centering
    \includegraphics[width=0.33\textwidth]{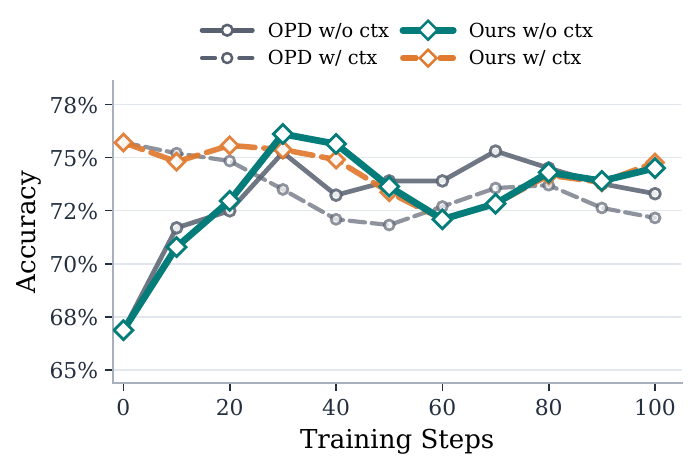}\hfill
    \includegraphics[width=0.33\textwidth]{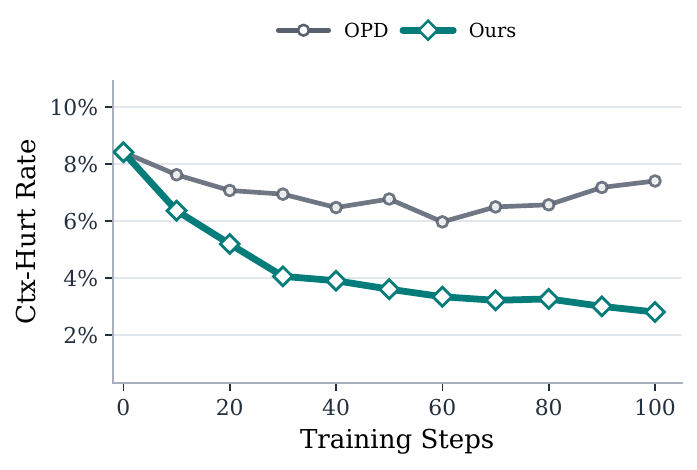}\hfill
    \includegraphics[width=0.33\textwidth]{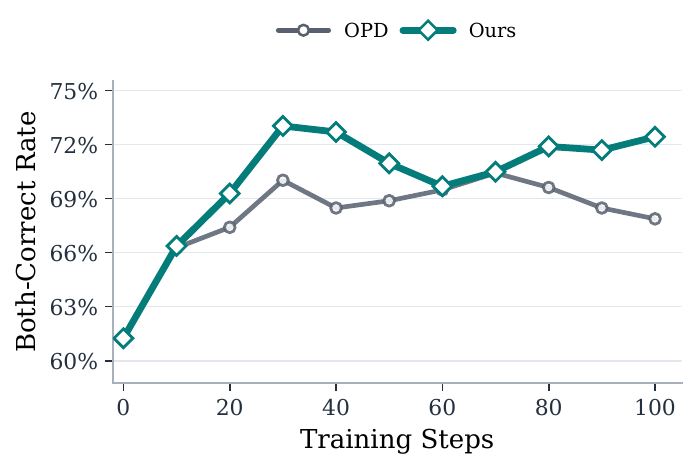}
    \caption{Qwen2.5-7B-Instruct, Safety}
  \end{subfigure}
  \caption{Regime~A settings (part 1). NCA consistently closes the accuracy gap and reduces harm rate.}
  \label{fig:apd_regime_a1}
\end{figure*}

\begin{figure*}[ht]
  \centering
  \begin{subfigure}[b]{\textwidth}
    \centering
    \includegraphics[width=0.33\textwidth]{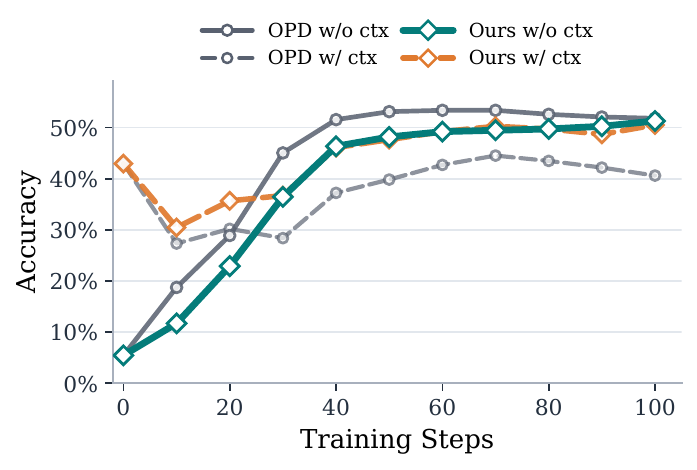}\hfill
    \includegraphics[width=0.33\textwidth]{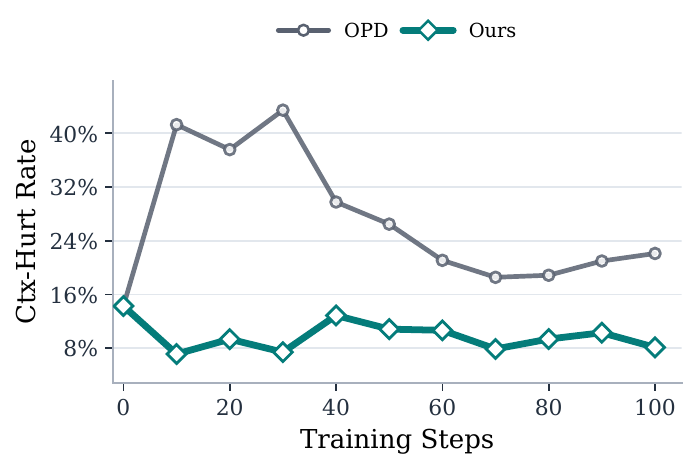}\hfill
    \includegraphics[width=0.33\textwidth]{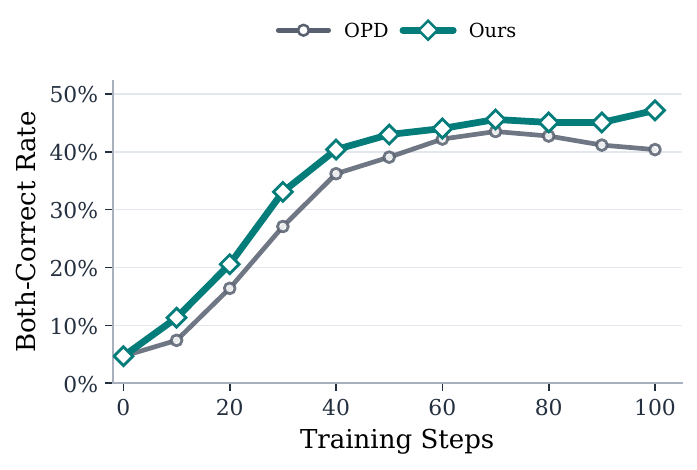}
    \caption{Qwen3-4B-Instruct-2507 (self), Sokoban}
  \end{subfigure}
  \\[1pt]
  \begin{subfigure}[b]{\textwidth}
    \centering
    \includegraphics[width=0.33\textwidth]{figures/paper/tasks/sokoban-q3-4b_to_1b7-ins_acc_intro.pdf}\hfill
    \includegraphics[width=0.33\textwidth]{figures/paper/tasks/sokoban-q3-4b_to_1b7-ins_context_hurts_no_context_correct_rate_intro.pdf}\hfill
    \includegraphics[width=0.33\textwidth]{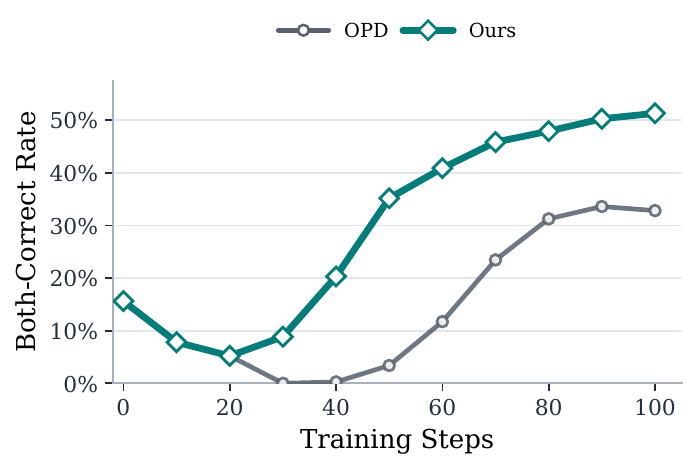}
    \caption{Qwen3-4B-Ins$\to$1.7B, Sokoban}
  \end{subfigure}
  \\[1pt]
  \begin{subfigure}[b]{\textwidth}
    \centering
    \includegraphics[width=0.33\textwidth]{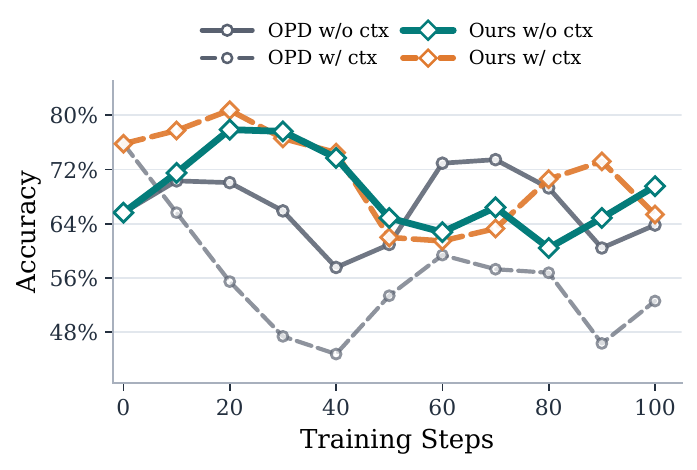}\hfill
    \includegraphics[width=0.33\textwidth]{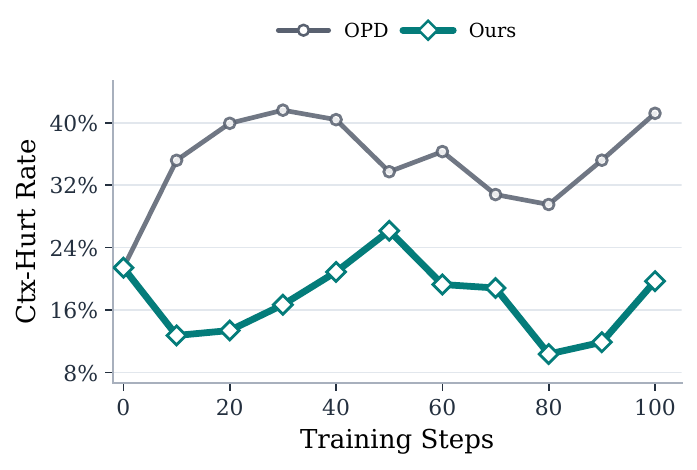}\hfill
    \includegraphics[width=0.33\textwidth]{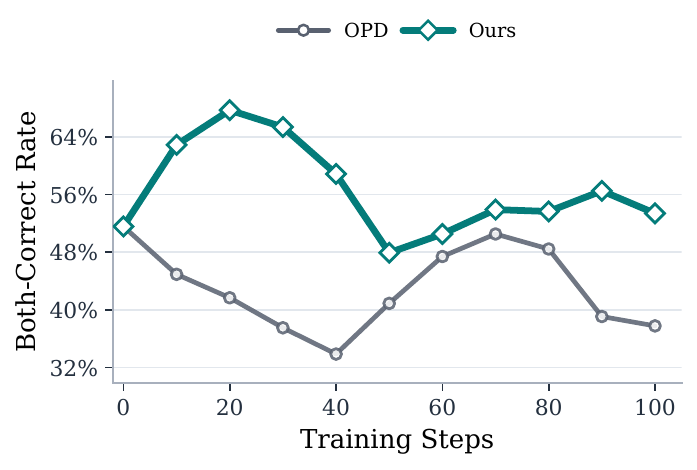}
    \caption{Qwen3-4B-Instruct-2507 (self), FrozenLake}
  \end{subfigure}
  \\[1pt]
  \begin{subfigure}[b]{\textwidth}
    \centering
    \includegraphics[width=0.33\textwidth]{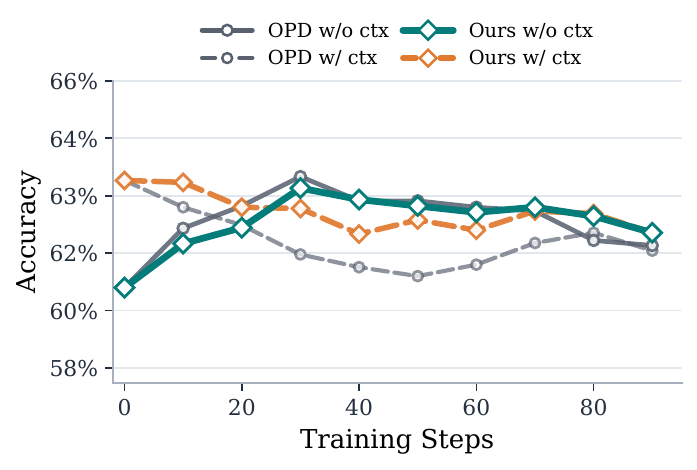}\hfill
    \includegraphics[width=0.33\textwidth]{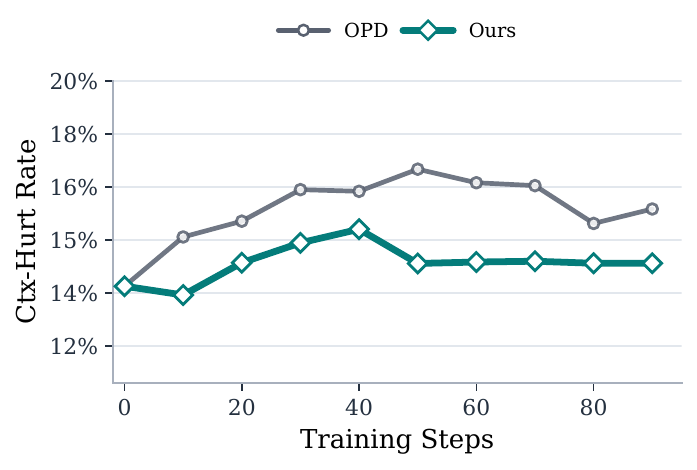}\hfill
    \includegraphics[width=0.33\textwidth]{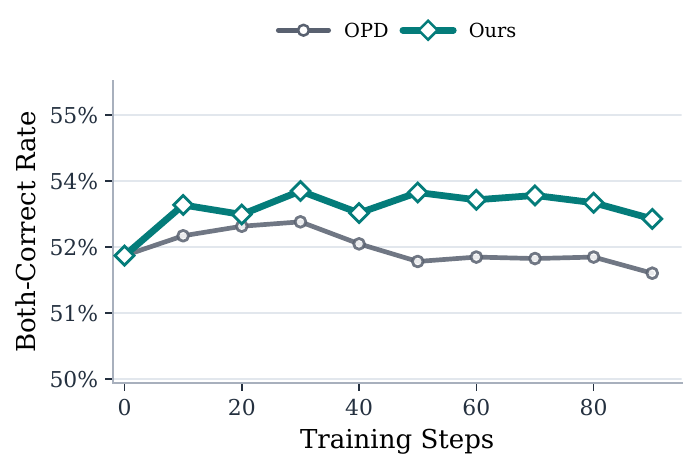}
    \caption{Qwen3-1.7B, Math}
  \end{subfigure}
  \caption{Regime~A settings (part 2). NCA consistently closes the accuracy gap and reduces harm rate.}
  \label{fig:apd_regime_a2}
\end{figure*}

\begin{figure*}[ht]
  \centering
  \begin{subfigure}[b]{\textwidth}
    \centering
    \includegraphics[width=0.33\textwidth]{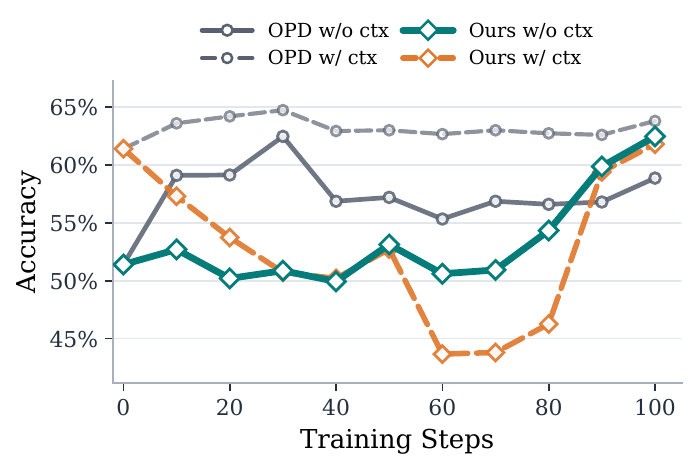}\hfill
    \includegraphics[width=0.33\textwidth]{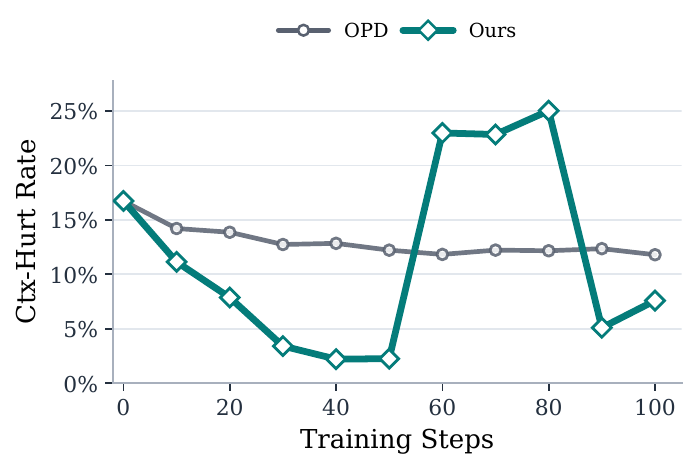}\hfill
    \includegraphics[width=0.33\textwidth]{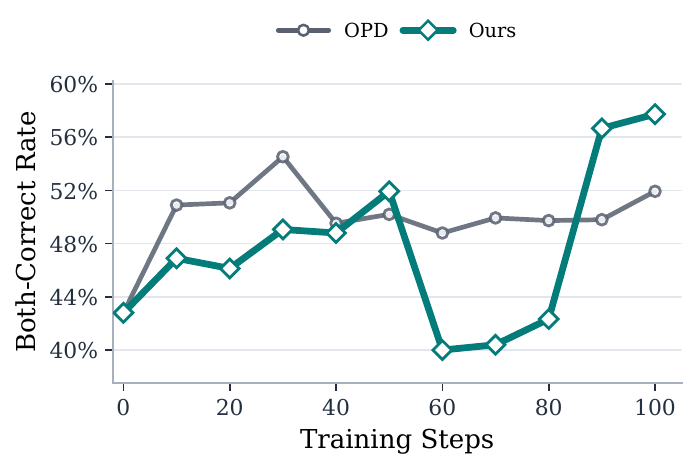}
    \caption{Qwen2.5-7B-Instruct, MedMCQA}
  \end{subfigure}
  \\[1pt]
  \begin{subfigure}[b]{\textwidth}
    \centering
    \includegraphics[width=0.33\textwidth]{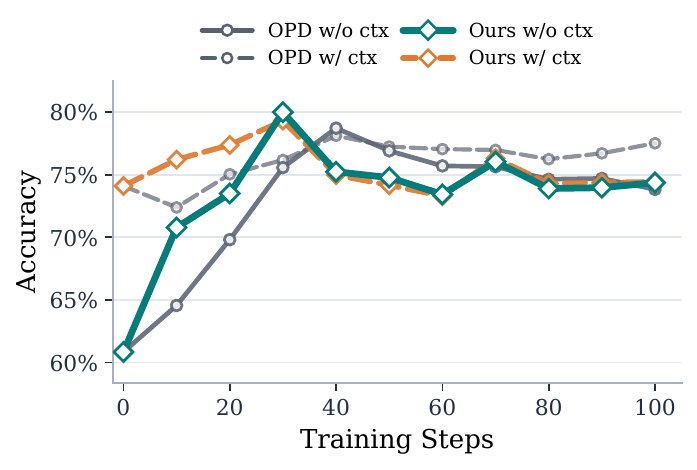}\hfill
    \includegraphics[width=0.33\textwidth]{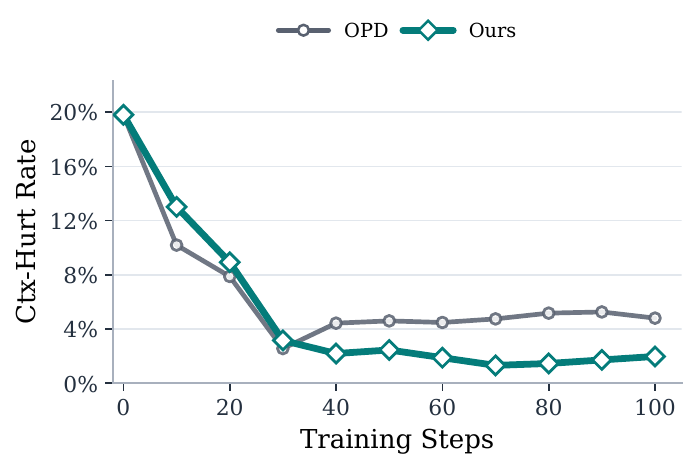}\hfill
    \includegraphics[width=0.33\textwidth]{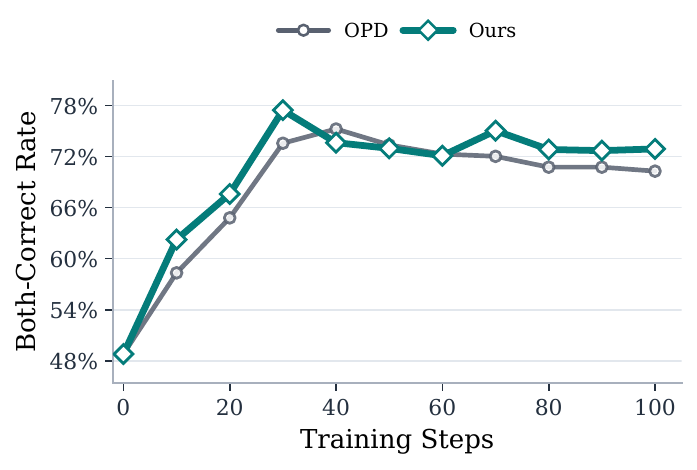}
    \caption{Llama-3.2-3B-Instruct, Safety}
  \end{subfigure}
  \\[1pt]
  \begin{subfigure}[b]{\textwidth}
    \centering
    \includegraphics[width=0.33\textwidth]{figures/paper/tasks/frozenlake-q3-4b_to_1b7-ins_acc_intro.pdf}\hfill
    \includegraphics[width=0.33\textwidth]{figures/paper/tasks/frozenlake-q3-4b_to_1b7-ins_context_hurts_no_context_correct_rate_intro.pdf}\hfill
    \includegraphics[width=0.33\textwidth]{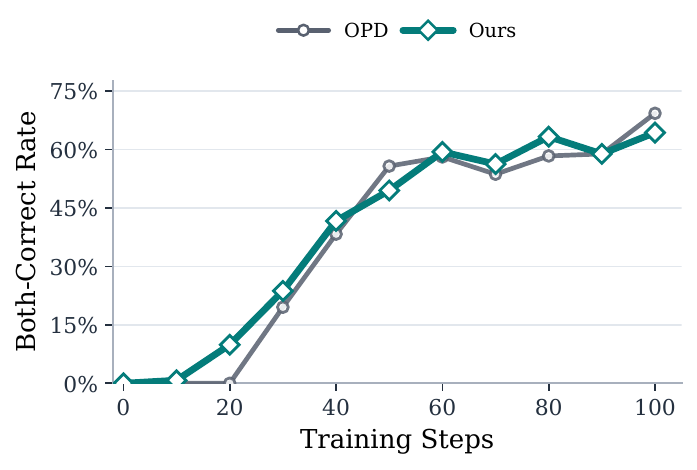}
    \caption{Qwen3-4B-Ins$\to$1.7B, FrozenLake}
  \end{subfigure}
  \\[1pt]
  \begin{subfigure}[b]{\textwidth}
    \centering
    \includegraphics[width=0.33\textwidth]{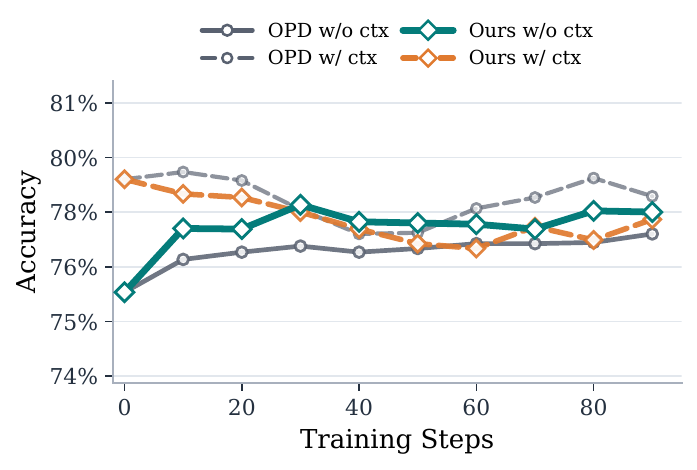}\hfill
    \includegraphics[width=0.33\textwidth]{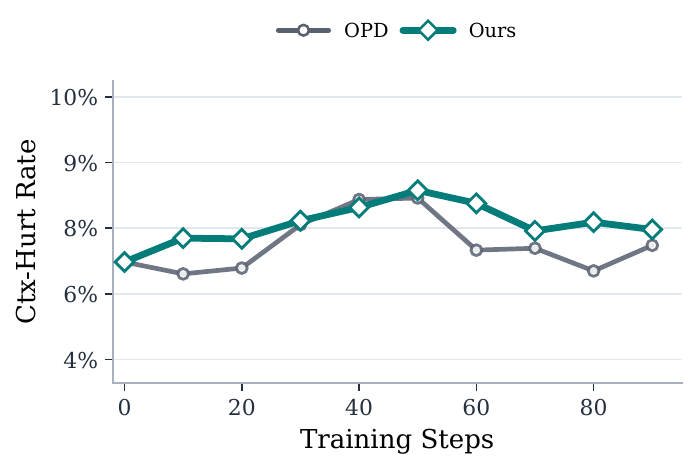}\hfill
    \includegraphics[width=0.33\textwidth]{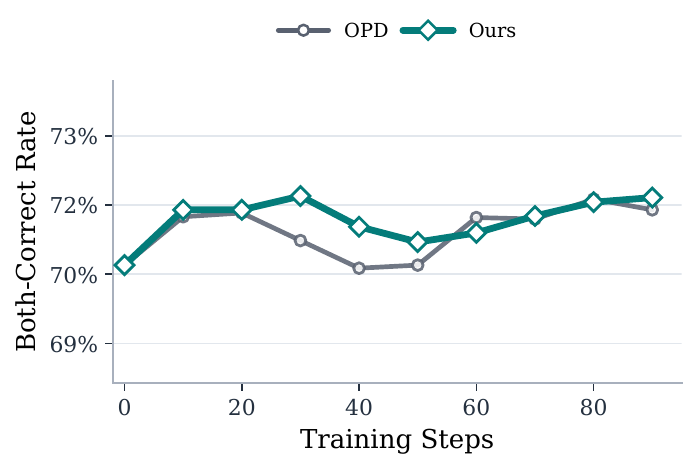}
    \caption{Qwen3-8B, Math}
  \end{subfigure}
  \caption{Regime~B settings. Context still provides useful scaffolding, and NCA trades some $\Acc_{x,c}$ for improved $\Acc_x$.}
  \label{fig:apd_regime_b}
\end{figure*}

\begin{figure*}[ht]
  \centering
  \begin{subfigure}[b]{\textwidth}
    \centering
    \includegraphics[width=0.33\textwidth]{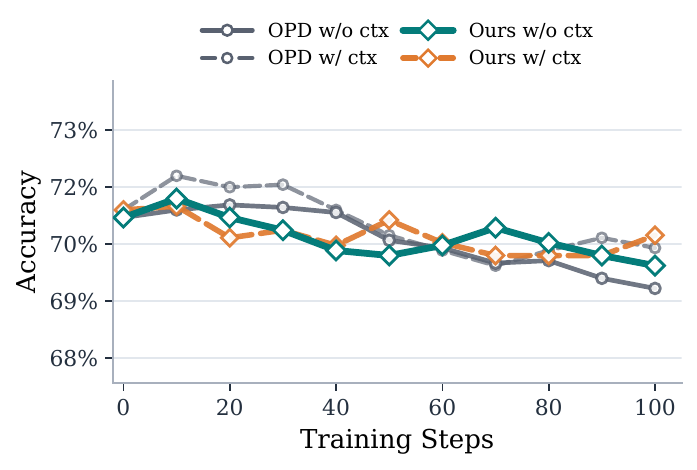}\hfill
    \includegraphics[width=0.33\textwidth]{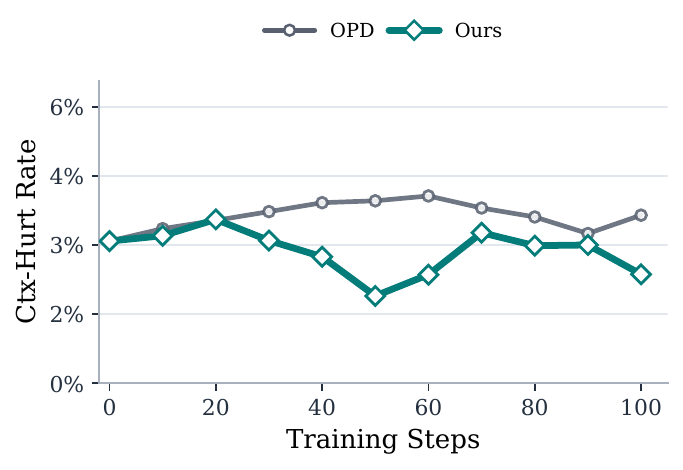}\hfill
    \includegraphics[width=0.33\textwidth]{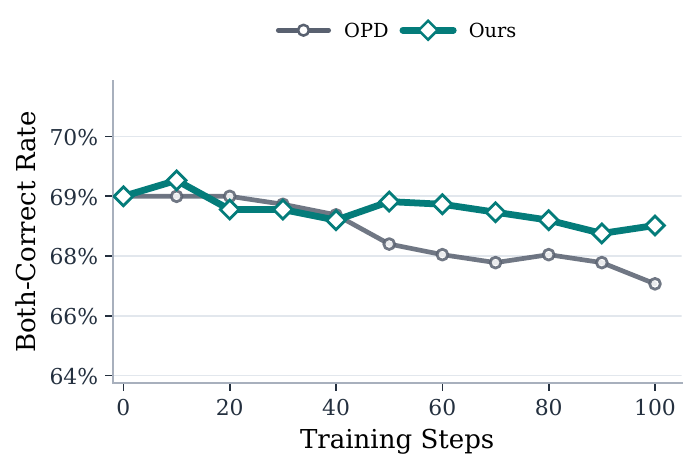}
    \caption{Qwen3-8B, MedMCQA}
  \end{subfigure}
  \\[1pt]
  \begin{subfigure}[b]{\textwidth}
    \centering
    \includegraphics[width=0.33\textwidth]{figures/paper/tasks/safety-q3-8b_acc_intro.pdf}\hfill
    \includegraphics[width=0.33\textwidth]{figures/paper/tasks/safety-q3-8b_context_hurts_no_context_correct_rate_intro.pdf}\hfill
    \includegraphics[width=0.33\textwidth]{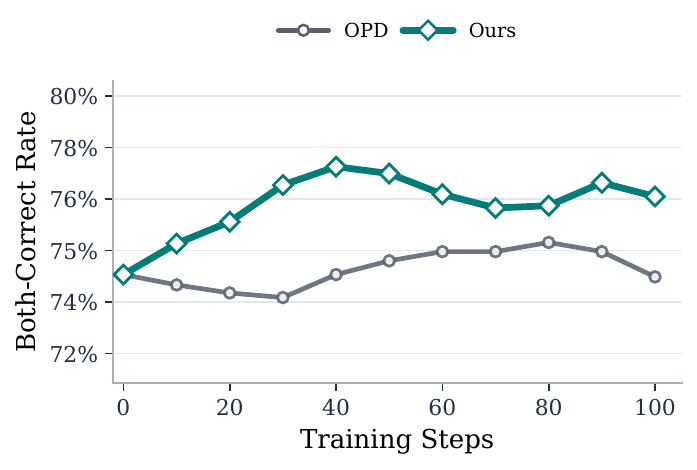}
    \caption{Qwen3-8B, Safety}
  \end{subfigure}
  \caption{Regime~C settings. Both views already agree, and NCA shows minimal additional effect.}
  \label{fig:apd_regime_c}
\end{figure*}

\end{document}